\documentclass[10pt,twocolumn,letterpaper]{article}

\usepackage[pagenumbers]{cvpr}

\usepackage{algorithm,algpseudocode}

\makeatletter
\renewcommand{\cvprsubsubsection}{\@startsection{subsubsection}{3}{\z@}%
  {-\baselineskip}{-.5em}{\tenbf}}
\renewcommand{\cvprsubsubsect}[1]{\cvprsubsubsection{\texorpdfstring{\hskip -1em.~}{}#1\texorpdfstring{.}{}}}
\renewcommand{\cvprssubsubsect}[1]{\cvprsubsubsection*{#1\texorpdfstring{.}{}}}
\makeatother

\usepackage{microtype}

\definecolor{cvprblue}{rgb}{0.21,0.49,0.74}
\usepackage[pagebackref,breaklinks,colorlinks,allcolors=cvprblue]{hyperref}

\def\confName{CVPR}
\def\confYear{2027}

\title{RecurNav: A Vision-Language Navigation Harness with Recursively Evolving Spatio-temporal Memory}

\author{%
Zhikai Wang\textsuperscript{2,*}\qquad Yuhang Peng\textsuperscript{3,*}\qquad Jingzhi Huang\textsuperscript{2}\qquad Yuhao Wu\textsuperscript{1}\qquad Yutong Du\textsuperscript{1}\\
Xiaoji Zheng\textsuperscript{4}\qquad Jiaru Zhong\textsuperscript{2}\qquad Yi Wang\textsuperscript{2}\qquad Jiangtao Gong\textsuperscript{1,\textdagger}\\
\textsuperscript{1}Shanghai Jiao Tong University\qquad \textsuperscript{2}The Hong Kong Polytechnic University\\
\textsuperscript{3}DeepRobotics\qquad \textsuperscript{4}Tsinghua University
}

\begin{document}
\raggedbottom
\maketitle
\begingroup
\renewcommand{\thefootnote}{*}
\footnotetext[1]{Equal contribution.\quad \textsuperscript{\textdagger} Corresponding author.}
\endgroup
\begin{abstract}
Reliable vision-language navigation requires tracking instruction progress during execution and using past failures to improve subsequent decisions. We present RecurNav, a training-free navigation harness that combines spatio-temporal route memory with recursively evolving experience memory for a frozen vision-language model (VLM). Within an attempt, route memory organizes task progress, visual history, and local geometry around the active subgoal. Together with measured motion, this memory enables the navigator to assess progress and correct its course during waypoint execution. Across attempts, recursive self-improvement (RSI) distills failed trajectories into conditional lessons and reviews them against the same trajectory observations and event records used for distillation. The harness retrieves relevant cases and lessons to guide verification of subgoal completion and stopping. Unresolved routes are retried with the updated memory, and their latest failures inform the next round of distillation and review. With GPT-6.1-sol, RecurNav achieves 76\% success on OpenNav-100 and 67\% on a 100-route RxR-CE subset, using 37\% fewer tokens than MIP Minimal on RxR-CE. With Qwen3.8-27B, experience-guided retries increase cumulative success on OpenNav-100 about 15\%.
\end{abstract}
    
\section{Introduction}
\label{sec:intro}

Vision-Language Navigation (VLN) requires an agent to follow natural-language instructions in unseen environments~\cite{anderson2018r2r,krantz2020continuous}. Many instructions depend on route progress. Given ``pass the first doorway on the left and enter the next,'' the two openings may look alike, yet the agent should walk past the first and turn into the second. Training-based navigators such as AwareVLN and Progress-Think learn to estimate this progress from trajectories~\cite{guo2026awarevln,wang2026progressthink}, but learned policies remain sensitive to the scenes and instruction styles seen in training~\cite{hong2025gsa,li2026slow4fast}. A training-free agent built on a frozen vision-language model (VLM) avoids retraining, yet it can judge progress only from the evidence its harness provides.

The route itself supplies such evidence: an earlier view of the first opening and the motion measured since then indicate whether the agent has passed it. Using this evidence raises two difficulties. First, the model must judge progress during a move. If its next view comes only after arrival, it detects a wrong turn too late. Second, a frozen model repeats its errors: an agent that once declared a doorway passed while the opening was still ahead will make the same claim on the next route.

We introduce \emph{RecurNav}, a training-free VLN harness that addresses both difficulties with memory (Figure~\ref{fig:recurnav-overview}). Within an attempt, it builds on the pose-indexed history of ST-VLN~\cite{peng2026remember} and the execution feedback of HarnessVLN~\cite{chen2026harnessvln}. For the active subgoal, the model inspects earlier views alongside its measured motion and selects a reachable waypoint, which a geometric controller tracks in world coordinates. Because the waypoint stays fixed, the harness can pause at checkpoints and show the model a fresh view, letting it continue or change course without losing the target. Advancing to the next subgoal is a separate decision that requires evidence from the route.

Across attempts, RecurNav uses memory-based recursive self-improvement (RSI)~\cite{yu2026recuris} to change what this decision requires. Building on reflection across trials~\cite{shinn2023reflexion} and reuse of navigation histories~\cite{dai2026evonav,tu2026retronav}, RSI turns failures into reusable checks. In the doorway example, the agent learns to look for evidence that it has moved beyond the opening. When the agent proposes completing a subgoal or stopping, a verifier retrieves relevant checks and tests them against the current route's views and measured motion. If evidence is missing, the verifier tells the agent what to check next, and later failures refine the checks.

\providecommand{\overviewwidth}{0.84\linewidth}
\begin{figure*}[t!]
  \centering
  \includegraphics[width=\overviewwidth]{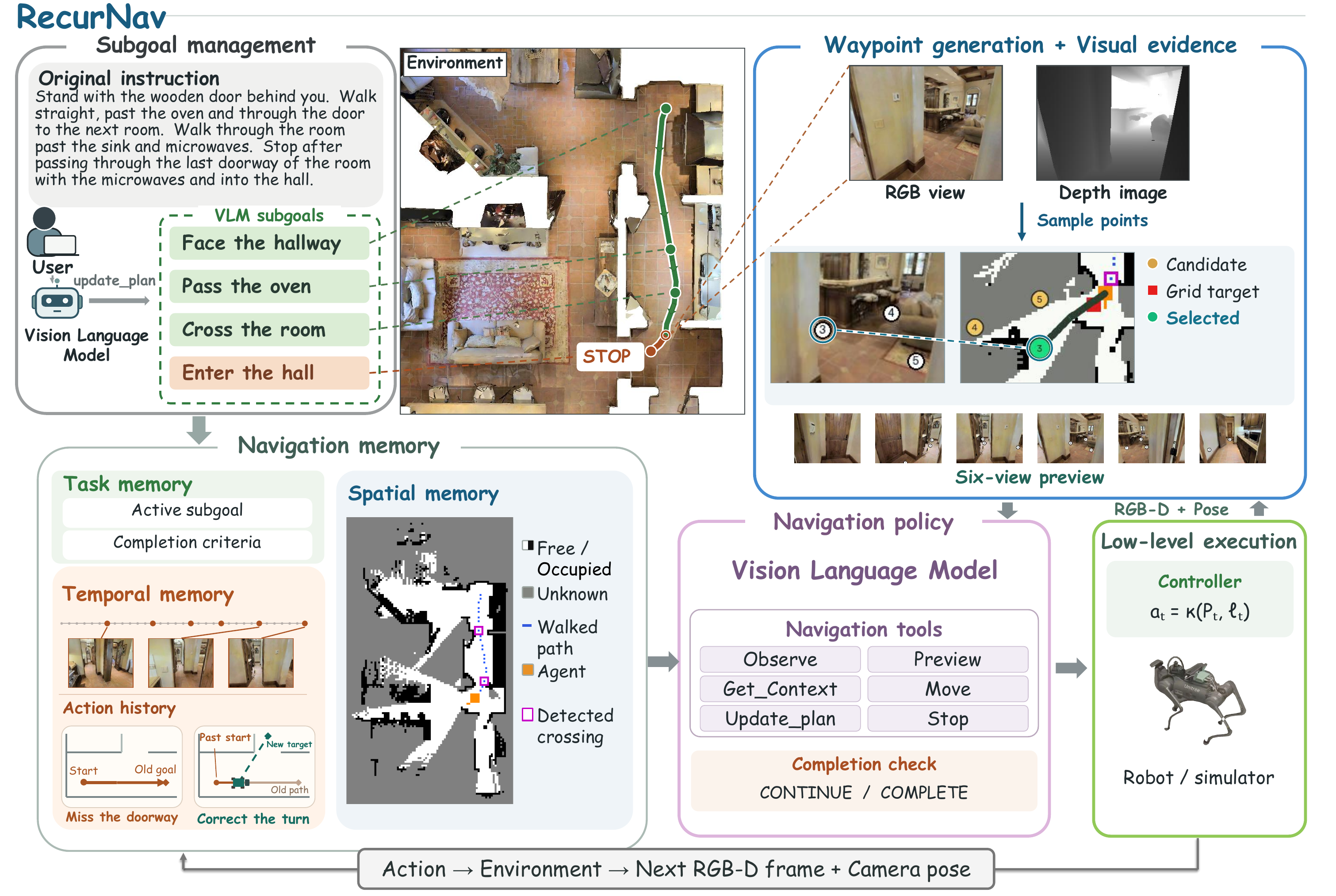}
  \caption{Overview of RecurNav. A frozen VLM divides the instruction into subgoals. Matching numbers in the camera view and local map identify the same reachable waypoint. The controller tracks the selected point in world coordinates, and orange dots mark past views retrieved from memory. In the doorway example, the agent recalls a move that missed the opening and selects a new waypoint.}
  \label{fig:recurnav-overview}
\end{figure*}

Our contributions are threefold.
\begin{itemize}
  \item \textbf{A training-free VLN harness.} RecurNav connects a frozen VLM's subgoal decisions to waypoint execution in continuous space, achieving 76\% mean SR on OpenNav-100 with GPT-6.1-sol.
  \item \textbf{Spatio-temporal memory for route progress.} Earlier views are selected by subgoal and distance traveled, and the active waypoint is held in world coordinates, so a new observation can correct the course during a move.
  \item \textbf{Recursively evolving experience memory.} Failed attempts revise the evidence required before accepting a subgoal or ending a route. With Qwen3.8-27B~\cite{qwen2026qwen3827b}, SR on OpenNav-100 rises from 58\% on the first attempt to 81\% after case retrieval and four RSI revisions.
\end{itemize}

\section{Related Work}
\label{sec:related}

\subsection{Vision-Language Navigation}

In continuous VLN, an agent must decide where to go and execute the motion described by instruction~\cite{anderson2018r2r,ku2020rxr,krantz2020continuous}. Training-based methods predict actions from egocentric video in NaVid~\cite{zhang2024navid}, an annotated map in MapNav~\cite{zhang2025mapnav}, or reasoning traces in FantasyVLN~\cite{zuo2026fantasyvln}, and AwareVLN and Progress-Think also learn to recognize how much of an instruction has been completed~\cite{guo2026awarevln,wang2026progressthink}. DualVLN, released through InternNav, separates the selection of waypoints from a learned motion policy~\cite{wei2026dualvln}, and LightNav-0 predicts a target point and generates a trajectory toward it~\cite{wang2026lightnav}. Once a target is chosen, however, the navigator still has to decide during execution whether the latest observation warrants continuing toward it.

Zero-shot methods let a pretrained model interpret the instruction while a planner determines how to move. InstructNav and CA-Nav guide motion with value maps~\cite{long2024instructnav,chen2025canav}, GC-VLN derives waypoints from spatial constraints in the instruction~\cite{yin2025gcvln}, and Open-Nav uses a language model to assess progress and select the next move~\cite{qiao2025opennav}. Agent harnesses further let the model decide when to observe and act: MIP Minimal lets it request a view or issue a sequence of actions~\cite{zhou2026mip}, AgenticNav exposes depth queries, visual recall, and pixel targets as tools~\cite{li2026agenticnav}, and HarnessVLN checks proposed actions against scene geometry and subgoal progress before dispatch~\cite{chen2026harnessvln}. RecurNav extends this control into execution, letting the model reconsider a waypoint while the controller is still tracking it.

Navigation evaluation also extends to unstructured robotic environments and interactive tasks. Bench2FreeAD benchmarks vision-based end-to-end navigation~\cite{peng2025bench2freead}, while FreeAskWorld supports closed-loop navigation in which agents actively request and interpret human guidance~\cite{peng2026freeaskworld}.

\subsection{Memory for VLN}

In object navigation, RefBeliefNav uses evidence uncertainty to guide additional observation and reconsider historical waypoints~\cite{huang2026refbeliefnav}. LERR-Nav combines execution summaries and map history with a learned policy that coordinates exploration, recovery, and reminiscing~\cite{huang2026lerrnav}.

SmartWay retains past waypoints for backtracking~\cite{shi2025smartway}, and ProFocus selects which waypoints to inspect again~\cite{xue2026profocus}. HSGM maps objects and candidate waypoints~\cite{li2026hsgm}, HAM-VLN records progress and failures on a world graph~\cite{liu2026hamvln}, and ST-VLN organizes instruction progress, pose-linked observation history, and accumulated geometry into task, temporal, and spatial memories~\cite{peng2026remember}. RecurNav instead organizes memory around the active subgoal: at a checkpoint, it compares earlier views of that subgoal with the current view and measured motion, while keeping the waypoint in world coordinates so the navigator can resume the move or choose a new target.

\subsection{Experience Reuse and Self-Improvement}

Whether experience helps on a new route depends on what the new instruction shares with earlier ones. Within a building, later instructions in a tour can refer to places already visited~\cite{krantz2023iterative}, and NavHarness retains maps and task records across sessions~\cite{zhao2026navharness}. Procedural experience can also transfer across buildings. Reflexion stores reflections on task feedback for later attempts~\cite{shinn2023reflexion}, and SMA distills procedures from verifier feedback on spatial reasoning tasks and uses later retrieval outcomes to estimate which procedures transfer reliably~\cite{zhang2026sma}. In navigation, EvoNav retrieves task and scene experience to anticipate actions and landmarks~\cite{dai2026evonav}, RetroNav reviews scene evidence to prepare a retry~\cite{tu2026retronav}, and slow4fast-VLN improves its fast policy with experience extracted from execution records~\cite{li2026slow4fast}.

RecurNav applies memory-based recursive self-improvement~\cite{yu2026recuris} to verification, revising what the verifier requires before it accepts a subgoal as complete or allows the agent to stop. A premature stop at one doorway yields a check for evidence of crossing the threshold. At another doorway, the verifier retrieves this check and looks for the crossing in the new route's views and measured motion. A later failure provides evidence for the next revision.

\section{Methodology}
\label{sec:method}

Given a route instruction, the agent navigates with RGB-D observations and camera poses and decides where to stop without access to the evaluation goal. RecurNav operates on two timescales (Figure~\ref{fig:recurnav-overview}). Within an attempt, the navigator reads and updates route memory. Across attempts, RSI revises the checks used to verify progress.

\subsection{Harness}
\label{sec:harness}

A frozen VLM chooses each tool call, and the harness maintains the route state, executes the call, and returns its outcome. Motion toward a selected waypoint is delegated to a geometric controller.

\subsubsection{Route state and decision cycle}
Let $k$ index model calls and $t_k$ the physical step observed at call $k$. The route state $S_k$ includes the instruction, task, temporal, and spatial memories, and the current execution status. Let $o_t$ denote the RGB-D observation and camera pose at step $t$. One decision cycle is
\begin{equation}
\begin{aligned}
 c_k &= \operatorname{Read}(S_k,o_{t_k}),\\
 u_k &\sim \pi_\theta(\,\cdot\mid c_k),\\
 (e_k,t_{k+1}) &= \operatorname{Execute}(S_k,u_k),\\
 S_{k+1} &= \operatorname{Update}(S_k,u_k,e_k).
\end{aligned}
\label{eq:decision-cycle}
\end{equation}
Here $c_k$ contains the active plan, selected views, recent tool exchanges, execution feedback, and available tools. The frozen policy $\pi_\theta$ outputs a tool name and arguments $u_k$, whose execution yields the evidence $e_k$ for the state update. A memory query leaves the physical step unchanged ($t_{k+1}=t_k$), whereas a waypoint move can advance several steps before the next model call. The evidence returned by the move includes the intervening observations and measured motion.

\subsubsection{Tool interface}
\texttt{Observe} returns the current camera view. In Habitat, \texttt{Preview} renders six headings at the same position and restores the original pose, so looking around costs no primitive actions (Table~\ref{tab:recurnav-tools}). \texttt{Get\_context} queries route memory without moving, \texttt{Move} hands a target to the controller, and \texttt{Update\_plan} and \texttt{Stop} submit proposals that take effect only after the verifier accepts them.

\begin{table}[!h]
\centering
\caption{RecurNav tool interface.}
\label{tab:recurnav-tools}
\footnotesize
\begin{tabular}{@{}p{0.26\linewidth}p{0.68\linewidth}@{}}
\toprule
Tool & Function \\
\midrule
\texttt{Observe} & Inspect the current frame and execution feedback.\\
\texttt{Preview} & View six directions with numbered waypoint marks.\\
\texttt{Get\_context} & Retrieve past events, evidence, or a selected frame.\\
\texttt{Update\_plan} & Submit ordered subgoals and completion evidence for review.\\
\texttt{Move} & Turn, move forward, or follow a waypoint until arrival, interruption, or review.\\
\texttt{Stop} & Propose a stop with a reason and supporting evidence.\\
\bottomrule
\end{tabular}
\end{table}

For each preview view $v$, the harness divides the image into five vertical strips. In each nonempty strip, it selects a floor pixel $s_i$ at the 80th percentile of candidate depths and keeps it if its depth $d_i$ is at least 0.8\,m. The candidate waypoint is placed 0.5\,m short of this pixel:
\begin{equation}
d_i'=\max(0.8,\,d_i-0.5),\qquad
\tilde q_i=F_t\!\left(B_{t,v}(s_i,d_i')\right),
\label{eq:waypoints}
\end{equation}
where $B_{t,v}$ back-projects the pixel at the shortened depth (meters) and $F_t$ sets its height to floor level. After removing near-duplicates, checking candidates against the local map, and filtering targets of failed approaches, the harness draws a numbered mark for each accepted target $q_i$ in the current view, and the VLM selects that ID with \texttt{Move}.

\subsection{Memory}
\label{sec:memory}

The three route memories share one reference, the frame $b$ at which the active subgoal began: task memory records $b$, temporal memory retrieves its view, and spatial memory expresses waypoint candidates relative to its camera heading. Progress is thus measured on the current subgoal rather than on the whole route.

\subsubsection{Task memory}
The harness stores the instruction and the ordered subgoals with their completion criteria, status, and supporting frame references. The active criterion states what the navigator must establish before advancing the plan, and each accepted transition sets $b$ for the next subgoal.

\subsubsection{Temporal memory}
Each physical step adds a view and its camera pose to the route history. Instead of the full history, each model call receives a few selected views: the view at $b$, the current view, a designated keyframe, and available progress evidence take priority, and spare slots are filled with views selected by distance traveled. The navigator can also query earlier events and designate a frame with \texttt{Get\_context}.

\subsubsection{Spatial memory}
RGB-D observations and camera poses update the local map used to validate waypoint candidates. To keep directions comparable as the camera turns, each candidate is expressed relative to the pose at $b$. If $u$ and $v$ are the rightward and forward components of the displacement from that pose to the candidate on the floor plane, its bearing is $(180/\pi)\operatorname{atan2}(u,v)$ degrees, with $0^\circ$ straight ahead and left negative.

\subsection{Memory-Guided Execution and Feedback}
\label{sec:review}
When the VLM selects a waypoint mark, the harness resolves its ID to the world target $q_i$. At each physical step $t$, spatial memory supplies a local tracking point $\ell_t$ on the route toward $q_i$, and the waypoint controller (Figure~\ref{fig:recurnav-overview}) computes
\begin{equation}
 a_t=\kappa(P_t,\ell_t),
 \label{eq:waypoint-controller}
\end{equation}
where $P_t$ is the agent's base pose (position and heading). In Habitat, $\kappa$ measures the horizontal distance and heading error to $\ell_t$. Outside the local arrival radius, it steps forward if the heading error is within one turn increment and otherwise turns toward $\ell_t$. Inside the radius, it signals arrival. A single waypoint request can therefore yield several primitive actions $a_t$.

Judging progress during a move requires knowing both how far the agent has walked and whether it has approached the target. Let $s$ be the physical step at selection, $x_j$ the camera position on the floor plane at step $j$, and $\bar q_i$ the waypoint on that plane. At step $t$, execution feedback reports the distance traveled and the signed gain toward the waypoint:
\begin{equation}
\begin{aligned}
 d_{s,t} &= \sum_{j=s+1}^{t}\|x_j-x_{j-1}\|_2,\\
 g_{s,t} &= \|x_s-\bar q_i\|_2-\|x_t-\bar q_i\|_2 .
\end{aligned}
\label{eq:gain}
\end{equation}
A positive $g_{s,t}$ means the agent is closer to the target than at selection. A checkpoint is due when travel since the previous review reaches the requested interval (default 1.5\,m, capped at 0.5\,m in the final stage). Outside the final stage, the controller can briefly defer review during a detour that increases the direct distance to the waypoint but shortens the remaining planned path.

At a checkpoint, the harness retains $q_i$ and returns a fresh view with $d_{s,t}$ and $g_{s,t}$. The VLM can inspect another view or retrieve an earlier frame, then resume $q_i$ or select a replacement. A collision or unreachable target ends the move: the harness releases $q_i$, records the failure and distance traveled, and requests a new preview for replanning. Reaching $q_i$ also closes the move. Neither outcome advances the plan: moving to the next subgoal requires an accepted \texttt{Update\_plan} call with frame evidence, and ending the episode requires a separately accepted \texttt{Stop} request.

\subsection{Recursively Evolving Memory}
\label{sec:evolving}
\begin{figure*}[t]
\centering
\includegraphics[width=\textwidth]{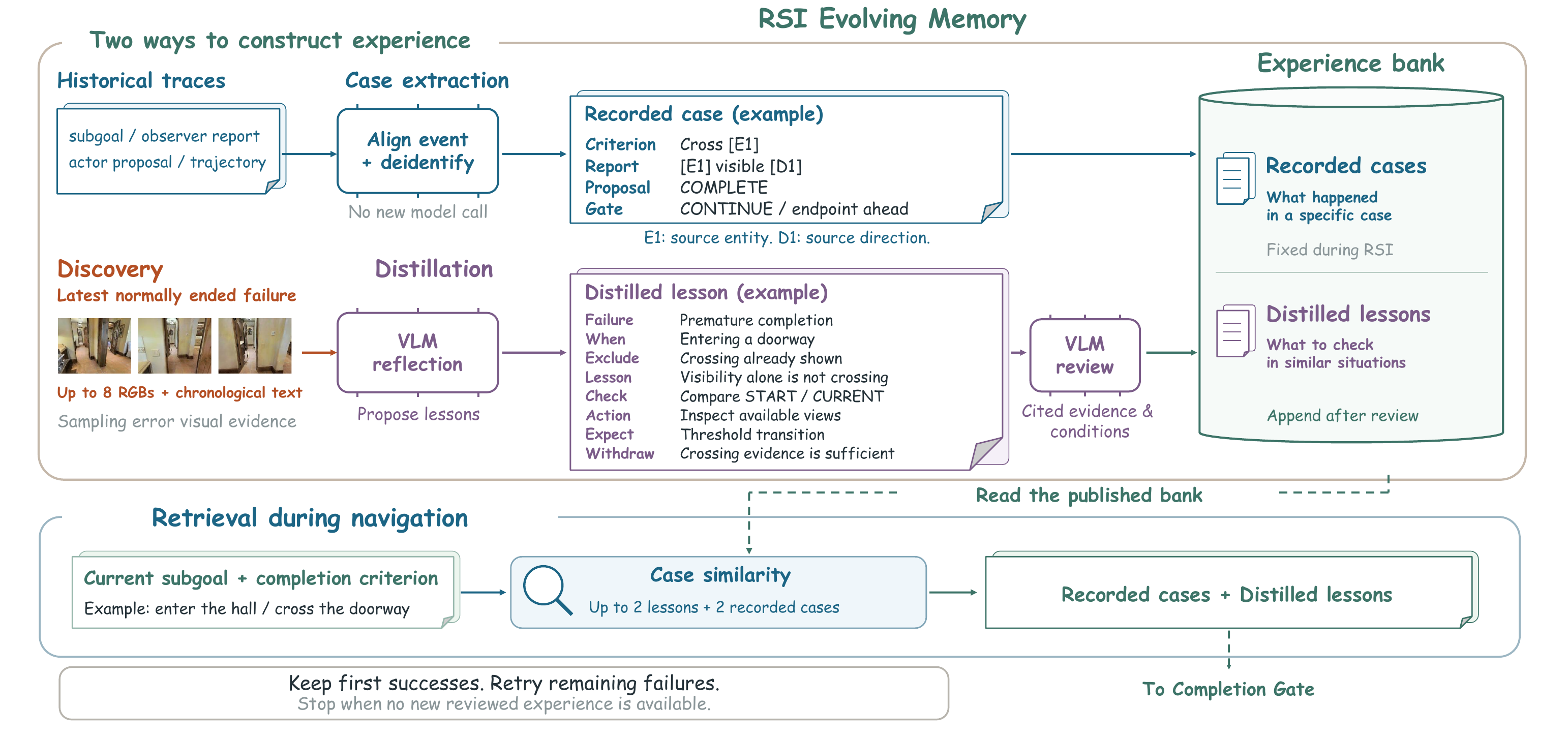}
\caption{Recursive self-improvement. RSI distills failed attempts into reusable checks. Before accepting a subgoal or stop, the verifier applies retrieved checks to the current views and measured motion.}
\label{fig:rsi-memory}
\end{figure*}

RSI improves subgoal-completion and stopping decisions by updating an experience bank while keeping model weights fixed (Figure~\ref{fig:rsi-memory}, Algorithm~\ref{alg:rsi}). The bank contains \emph{recorded cases} $C$, which preserve specific historical decisions, and \emph{distilled lessons} $E$, which describe conditional checks for similar situations. Both are retrieved during navigation to guide the completion gate. To generate new lessons, RSI constructs \emph{trajectory evidence} $x$ from each failed attempt: chronological event records and sampled images used for distillation and review. Thus, $x$ supplies the evidence for constructing lessons, while $C$ and $E$ supply reusable guidance for navigation.

\begin{algorithm}[!t]
\caption{Recursive Self-Improvement (RSI)}
\label{alg:rsi}
\small
\begin{algorithmic}[1]
\State \textbf{Input:} Failed traces $F$, cases $C$, lessons $E$, round limit $K$
\State \textbf{Output:} Updated lessons $E$, remaining failures $F$
\State $r \gets 0$
\While{$F \neq \emptyset \land r < K$}
    \State $A \gets \emptyset$
    \ForAll{$\tau \in F$}
        \State $x \gets \Call{Read}{\tau}$
        \State $L \gets \Call{Distill}{x}$
        \State $A \gets A \cup \Call{Review}{L,x}$
    \EndFor
    \State $\Delta \gets A \setminus E$
    \If{$\Delta = \emptyset$}
        \State \textbf{break}
    \EndIf
    \State $E \gets E \cup \Delta$
    \State $F \gets \Call{Retry}{F,C,E}$
    \State $r \gets r+1$
\EndWhile
\State \Return $(E,F)$
\end{algorithmic}
\end{algorithm}

\subsubsection{Discovery}
For each normally terminated but unsuccessful attempt $\tau\in F$, \textsc{Read} constructs trajectory evidence $x$ by aligning proposed decisions with the active subgoal, the verifier's verdicts, and the visual and motion evidence available at each step. This evidence covers the failed attempt across time, with up to 64 chronological entries and eight RGB views linked by frame identifiers. Sampling prioritizes the first and last views and frames where a subgoal begins or its completion is assessed, then fills the remaining image slots uniformly. Distillation and review receive the same evidence $x$. Without images, lessons must be supported by the text records.

\subsubsection{Distillation}
\textbf{Recorded cases.} Before the RSI loop, historical traces are converted into recorded cases $C$ without a new model call. Each case summarizes one decision event for later retrieval. Each case aligns a subgoal criterion, observer report, agent proposal and its rationale, and completion-gate verdict at the same decision event. In Figure~\ref{fig:rsi-memory}, \emph{Criterion} states the required transition; \emph{Report} records the available observation; \emph{Proposal} is the agent's COMPLETE/CONTINUE judgment; and \emph{Gate} records the verifier's verdict and reason. Thus, the example preserves a disagreement: the agent claims completion while the gate reports that the endpoint remains ahead. De-identification replaces source terms with local symbols such as \texttt{[E1]} for an entity term and \texttt{[D1]} for a direction, using the same symbol for repeated terms within a case. Cases also retain whether the report was produced at the decision frame or carried over from an earlier frame.

\textbf{Distilled lessons.} For each failed attempt, \textsc{Distill} uses its trajectory evidence $x$ to ask a VLM for up to two candidate lessons $L$. The eight attributes shown in Figure~\ref{fig:rsi-memory} specify the scope of a lesson and how to use it. \emph{Failure} identifies the local failure pattern, such as premature completion. \emph{When} specifies the subgoal and conditions where the advice applies, while \emph{Exclude} identifies exceptions that make it inapplicable. \emph{Lesson} states the transferable insight. \emph{Check} describes how to examine observations or temporal evidence, and \emph{Expect} describes the observable finding that would support the judgment. \emph{Action} gives a conditional response to the check, including how to proceed when evidence is insufficient. Finally, \emph{Withdraw} specifies when to stop applying the advice in the current situation; the lesson remains in the bank for future use.

For the doorway example, visibility alone does not establish crossing. The check compares the subgoal's start view with the current view, seeking evidence of a transition across the threshold. If this evidence is missing, the suggested action is to inspect available views; once crossing is sufficiently supported, the additional check can cease. \emph{Exclude} governs whether the advice applies when consulted, whereas \emph{Withdraw} governs when its use ends.

\textbf{Evidence review.} A separate VLM call, \textsc{Review}, receives the candidate lessons $L$ together with the same trajectory evidence $x$ used for distillation, including the chronological records and available RGB images. It judges whether the cited evidence supports each lesson and its conditions of use. The reviewer returns a support decision, a reason, and evidence identifiers for each candidate. The code checks that these identifiers exist in $x$ and that each approved review shares at least one cited identifier with its candidate. Supported, valid candidates are added to $A$.

\subsubsection{Retrieval}
During navigation, a proposed subgoal completion or stop triggers retrieval from the case collection $C$ and lesson collection $E$ before the completion gate. The query combines the active subgoal and its completion criterion. We extract \emph{obligations}, task requirements such as crossing, turning, passing, or following a temporal order, from these texts. Both types of memory are ranked by the number of shared obligations plus lexical Jaccard similarity. For lessons, lexical matching uses the \emph{When} and \emph{Check} fields; for cases, it uses the de-identified criterion's structural terms. Case selection also matches observation-report freshness and keeps at most one case per source trajectory.

Up to two lessons and two cases share a 3{,}600-character context budget, with lessons placed first. Their fields enter the completion gate as textual guidance: cases illustrate earlier proposal--verdict pairs, while lessons specify applicable checks and expected evidence. The gate combines this guidance with the current visual and motion evidence to assess the proposal. In the doorway example, the retrieved lesson directs attention to the transition between the start and current views before completion is accepted.

\subsubsection{Update}
After all failed attempts have been reviewed, $A$ contains the approved candidate lessons for the current round. The harness appends $\Delta=A\setminus E$ to the lesson collection $E$, identifying duplicates by the content of all eight lesson fields. The case collection $C$ stays fixed, and existing lessons are retained. Each unresolved route restarts at its original pose using the updated bank. Successful routes retain their first successful outcomes and leave $F$. For each episode that fails again, the next RSI round analyzes the trace from its latest retry. When available, advice actually submitted during that attempt is included in the next distillation input, allowing the VLM to refine an earlier lesson's conditions or exclusions in light of the new failure. This closes the recursive loop between experience, navigation, and subsequent reflection. The loop stops when $F$ is empty, review yields no new lesson, or $K$ rounds have run.

\section{Experiments}
\label{sec:experiments}

\begin{table*}[!t]
\centering
\caption{Navigation results on R2R-CE val-unseen. OpenNav-100~\cite{qiao2025opennav} and RecurSet-200 are subsets of this split.}
\label{tab:navigation}
\footnotesize
\renewcommand{\arraystretch}{0.99}
\setlength{\fboxsep}{1.2pt}
\setlength{\tabcolsep}{2pt}
\begin{tabular*}{\textwidth}{@{\extracolsep{\fill}}lllrrrrrr@{}}
\toprule
\textbf{Method} & \textbf{Backbone} & \textbf{Camera views} & SR$\uparrow$ & OSR$\uparrow$ & SPL$\uparrow$ & nDTW$\uparrow$ & NE (m)$\downarrow$ & Steps$\downarrow$\\
\midrule
\multicolumn{9}{@{}l@{}}{\makebox[0pt][l]{\colorbox[gray]{0.94}{\makebox[\dimexpr\linewidth-2\fboxsep\relax][l]{\strut\itshape Training-based}}}\strut}\\
AwareVLN$_r$~\cite{guo2026awarevln} & AwareVLN (4-bit) & Single view & 65.2 & 76.0 & 56.9 & 66.0 & 3.89 & 110\\
InternNav$_r$~\cite{wei2026dualvln} & InternVLA & Single view & 58.5 & 65.9 & 53.8 & -- & 4.70 & 99\\
LightNav$_r$~\cite{wang2026lightnav} & LightNav & Single view & 68.0 & 72.2 & 62.5 & 63.3 & 3.99 & 84\\
NaVid~\cite{zhang2024navid} & NaVid & Single view & 37.4 & 49.1 & 35.9 & -- & 5.47 & --\\
MapNav~\cite{zhang2025mapnav} & MapNav & Single view & 39.7 & 53.0 & 37.2 & -- & 5.43 & --\\
\addlinespace[2pt]
\multicolumn{9}{@{}l@{}}{\makebox[0pt][l]{\colorbox[gray]{0.94}{\makebox[\dimexpr\linewidth-2\fboxsep\relax][l]{\strut\itshape Training-based\normalfont\enspace\textperiodcentered\enspace RecurSet-200}}}\strut}\\
AwareVLN$_r$~\cite{guo2026awarevln} & AwareVLN (4-bit) & Single view & 61.0 & 76.0 & 52.67 & 63.81 & 4.03 & 118.56\\
AwareVLN$_r$~\cite{guo2026awarevln} & AwareVLN (FP16) & Single view & 62.5 & 75.5 & 53.81 & 64.86 & 3.83 & 110.25\\
InternNav$_r$~\cite{wei2026dualvln} & InternVLA & Single view & 58.5 & 68.0 & 54.00 & -- & 4.55 & 91.80\\
LightNav$_r$~\cite{wang2026lightnav} & LightNav & Single view & 64.5 & 71.5 & 59.27 & 61.87 & 4.13 & 83.30\\
\midrule
\multicolumn{9}{@{}l@{}}{\makebox[0pt][l]{\colorbox[gray]{0.94}{\makebox[\dimexpr\linewidth-2\fboxsep\relax][l]{\strut\itshape Training-free}}}\strut}\\
CA-Nav + temporal memory$_r$~\cite{chen2025canav,peng2026remember} & Perception stack & Single view & 24.7 & 42.3 & 12.6 & 22.5 & 7.57 & 184\\
GC-VLN$_r$~\cite{yin2025gcvln} & DeepSeek-R1 & Single view & 14.1 & 28.9 & 6.6 & -- & 8.31 & 366\\
InstructNav~\cite{long2024instructnav} & GPT-4 + GPT-4V & 6 views & 31.0 & -- & 24.0 & -- & 6.89 & --\\
HSGM$_r$~\cite{li2026hsgm} & Qwen3-VL-8B & 3 views & 4.9 & 12.0 & 4.8 & 27.2 & 8.79 & 34\\
HSGM$_r$~\cite{li2026hsgm} & GPT-5 & 3 views & 46.3 & 56.9 & 32.4 & 46.3 & 5.47 & 43\\
SmartWay$_r$~\cite{shi2025smartway} & Qwen3-VL-8B & Panoramic & 17.8 & 36.1 & 13.4 & 38.0 & 7.39 & 12\\
Open-Nav$_r$~\cite{qiao2025opennav} & Qwen3-8B-FP8 & Panoramic & 10.7 & 14.8 & 9.3 & 43.0 & 7.87 & 6\\
HAM-VLN~\cite{liu2026hamvln} & Gemini-3.1-Pro & Panoramic & 61.0 & 78.7 & 48.1 & -- & 3.92 & --\\
\addlinespace[2pt]
\multicolumn{9}{@{}l@{}}{\makebox[0pt][l]{\colorbox[gray]{0.94}{\makebox[\dimexpr\linewidth-2\fboxsep\relax][l]{\strut\itshape Training-free\normalfont\enspace\textperiodcentered\enspace OpenNav-100}}}\strut}\\
SmartWay~\cite{shi2025smartway} & GPT-4o & Panoramic & 29.0 & 51.0 & 22.46 & -- & 7.01 & --\\
Open-Nav~\cite{qiao2025opennav} & GPT-4 & Panoramic & 19.0 & 23.0 & 16.10 & 45.79 & 6.70 & --\\
AgenticNav~\cite{li2026agenticnav} & Gemini-3.7-Flash & Panoramic & 76.0 & 78.0 & 66.50 & 69.38 & 3.48 & --\\
EvoNav~\cite{dai2026evonav} & Gemini-2.5-Pro & Panoramic & 43.0 & 51.0 & 37.77 & 62.38 & 5.04 & --\\
\addlinespace[3pt]
HarnessVLN$_r$~\cite{chen2026harnessvln} & GPT-6.1-sol & Panoramic & 59.0 & \underline{74.0} & 43.14 & 63.51 & \underline{3.88} & 191.37\\
MIP Minimal$_r$~\cite{zhou2026mip} & GPT-6.1-sol & Single view & \underline{68.0} & 72.0 & \underline{59.78} & \underline{68.62} & 4.29 & \underline{93.94}\\
\textbf{RecurNav (Ours)} & GPT-6.1-sol & Single + Preview & \textbf{76.0} & \textbf{79.0} & \textbf{69.70} & \textbf{78.00} & \textbf{3.16} & \textbf{70.02}\\
RecurNav (Ours) & Qwen3.8-27B & Single + Preview & 58.0 & 66.0 & 49.19 & 61.75 & 5.15 & 98.41\\
RecurNav (Ours)$^{*}$ & Qwen3.8-27B & Single + Preview & \textbf{81.0} & \textbf{85.0} & 62.29 & 63.96 & \textbf{2.96} & 119.73\\
\addlinespace[2pt]
\multicolumn{9}{@{}l@{}}{\makebox[0pt][l]{\colorbox[gray]{0.94}{\makebox[\dimexpr\linewidth-2\fboxsep\relax][l]{\strut\itshape Training-free\normalfont\enspace\textperiodcentered\enspace RecurSet-200}}}\strut}\\
CA-Nav$_r$~\cite{chen2025canav} & GPT-4 & Single view & 25.5 & 53.0 & 10.98 & 17.16 & 7.08 & 213.59\\
CA-Nav$_r$~\cite{chen2025canav} & Qwen3-VL-8B & Single view & 24.0 & 50.5 & 10.58 & 15.92 & 7.12 & 220.42\\
SmartWay$_r$~\cite{shi2025smartway} & Qwen3-VL-8B & Panoramic & 17.0 & 35.5 & 12.17 & 40.72 & 6.91 & 11.56\\
Open-Nav$_r$~\cite{qiao2025opennav} & Qwen3-8B-FP8 & Panoramic & 11.0 & 15.5 & 10.21 & 44.20 & 7.41 & 6.07\\
MIP Waypoint$_r$~\cite{zhou2026mip} & Qwen3.8-27B & 4 views & 36.83 & 59.67 & 28.09 & 47.26 & 5.69 & 60.95\\
RecurNav (Ours) & Qwen3-VL-8B & Single + Preview & 7.50 & 36.00 & 4.99 & 19.54 & 9.61 & 192\\
RecurNav (Ours) & Qwen3.8-27B & Single + Preview & 44.75 & 55.88 & 38.17 & 55.49 & 5.18 & 108.54\\
RecurNav (Ours)$^{*}$ & Qwen3.8-27B & Single + Preview & \textbf{72.5} & \textbf{85.0} & \textbf{58.47} & \textbf{63.14} & \textbf{3.38} & 127.44\\
\bottomrule
\end{tabular*}
\par\smallskip
{\raggedright\scriptsize RecurNav receives one current view and can request a six-direction preview. $r$ marks reproductions and $^{*}$ marks cumulative RSI results.\par}
\end{table*}

\subsection{Experimental Setup}
\begin{figure}[!htbp]
\centering
\includegraphics[width=\columnwidth]{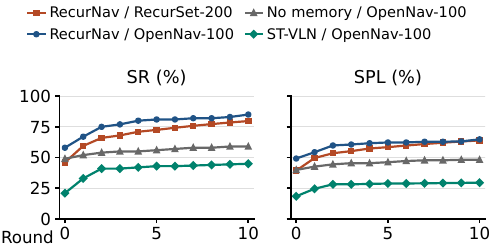}
\caption{Cumulative SR and SPL over retry rounds. RecurNav and the no-memory control use Qwen3.8-27B. The control uses its independently rerun baseline (49\% initial SR). ST-VLN uses Qwen3-VL-8B with our RSI procedure.}
\label{fig:rsi-rounds}
\end{figure}

\subsubsection{Benchmarks}
Following MIP~\cite{zhou2026mip}, we evaluate on OpenNav-100, the 100-episode R2R-CE val-unseen subset released with Open-Nav~\cite{qiao2025opennav}, and on 100 RxR-CE val-unseen routes with guide instructions~\cite{krantz2020continuous,ku2020rxr}. We report SR, OSR, SPL, nDTW, NE, and primitive steps. An episode succeeds if the agent stops within 3\,m of the goal. RecurNav, MIP Minimal, and HarnessVLN~\cite{chen2026harnessvln} use GPT-6.1-sol, identical starting poses, and a budget of 250 primitive actions per route on both sets. We access GPT-6.1-sol via API and report the mean of two RecurNav runs on OpenNav-100. For local Qwen3.8-27B inference, we run one BF16 vLLM replica per NVIDIA A800-SXM4 GPU (80\,GB) with a 16,384-token context limit.

\subsubsection{RecurSet-200}
To analyze failure modes, we curate 200 instructions from R2R-CE val-unseen, 50 each from the doorway, walking, stairs, and abstract stopping categories.

\subsubsection{Retries}
Each round keeps successes and restarts unresolved routes. Round~1 adds historical cases, and later rounds add RSI revisions. The no-memory control executes the same ordered episode cohorts, including episodes it has already solved. Starred rows in Table~\ref{tab:navigation} report cumulative results after case retrieval and four revisions. Figure~\ref{fig:rsi-rounds} extends the comparison to round ten.

\begin{table}[!htbp]
\centering
\caption{Navigation results on 100 RxR-CE val-unseen routes.}
\label{tab:navigation-rxr}
\footnotesize
\renewcommand{\arraystretch}{1.10}
\setlength{\fboxsep}{1.2pt}
\setlength{\tabcolsep}{1.4pt}
\begin{tabular*}{\linewidth}{@{\extracolsep{\fill}}lrrrrr@{}}
\toprule
\textbf{Method} & SR$\uparrow$ & OSR$\uparrow$ & SPL$\uparrow$ & nDTW$\uparrow$ & NE (m)$\downarrow$\\
\midrule
\multicolumn{6}{@{}l@{}}{\makebox[0pt][l]{\colorbox[gray]{0.94}{\makebox[\dimexpr\linewidth-2\fboxsep\relax][l]{\strut\itshape Training-based}}}\strut}\\
MapNav~\cite{zhang2025mapnav} & 32.6 & -- & 27.7 & 43.5 & 7.62\\
\midrule
\multicolumn{6}{@{}l@{}}{\makebox[0pt][l]{\colorbox[gray]{0.94}{\makebox[\dimexpr\linewidth-2\fboxsep\relax][l]{\strut\itshape Training-free}}}\strut}\\
HAM-VLN~\cite{liu2026hamvln} & 52.7 & -- & 35.7 & 54.3 & 6.27\\
\addlinespace[4pt]
HarnessVLN$_r$~\cite{chen2026harnessvln} & 38.0 & 44.0 & 27.04 & 56.91 & 7.05\\
MIP Minimal$_r$~\cite{zhou2026mip} & \underline{66.0} & \textbf{70.0} & \underline{53.84} & \underline{68.10} & \textbf{4.32}\\
\textbf{RecurNav (Ours)} & \textbf{67.0} & \underline{69.0} & \textbf{55.70} & \textbf{70.34} & \underline{4.39}\\
\bottomrule
\end{tabular*}
\end{table}

\begin{table}[!htbp]
\centering
\caption{Experience transfer. Gemini-3.8-Flash-Low on 100 R2R-CE val-unseen episodes outside OpenNav-100, with and without experience collected by Qwen3.8-27B on OpenNav-100.}
\label{tab:experience-transfer}
\footnotesize
\begin{tabular*}{\linewidth}{@{\extracolsep{\fill}}lrr@{}}
\toprule
Metric & No experience & OpenNav-100 experience \\
\midrule
SR (\%) $\uparrow$ & 57.00 & \textbf{60.00} \\
SPL (\%) $\uparrow$ & 51.61 & \textbf{53.11} \\
OSR (\%) $\uparrow$ & 64.00 & 70.10 \\
\bottomrule
\end{tabular*}
\end{table}

\subsection{Main Results}
\label{sec:main-results}
With GPT-6.1-sol, RecurNav achieves 76\% SR on OpenNav-100, 8 points above MIP Minimal and 17 above HarnessVLN under the same backbone and budget, while taking the fewest steps (Table~\ref{tab:navigation}). On RxR-CE, RecurNav achieves 67\% SR, close to MIP Minimal, and leads the comparison in SPL and nDTW (Table~\ref{tab:navigation-rxr}).

\begin{figure}[!htbp]
\centering
\IfFileExists{figures/rxr-routes.pdf}{\includegraphics[width=.84\linewidth]{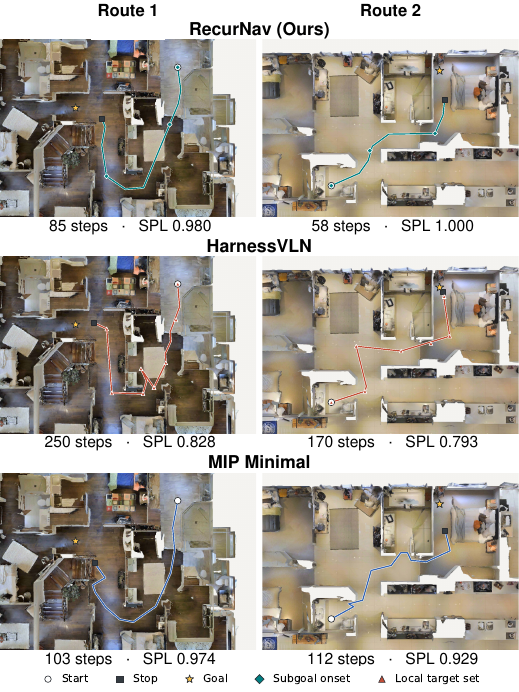}}{\includegraphics[width=.84\linewidth]{figures/rxr-routes.pdf}}
\caption{Trajectories on two RxR-CE routes. Step counts reveal turning costs that SPL ignores.}
\label{fig:rxr-routes}
\end{figure}

On the first route in Figure~\ref{fig:rxr-routes}, RecurNav and MIP Minimal follow similarly efficient paths, with RecurNav reaching the goal in 18 fewer steps. Turns in place count as steps and leave the path length unchanged, so SPL obscures that difference. On the second route, RecurNav also travels a shorter distance. It reaches the goal in 58 steps, roughly half as many as MIP Minimal and a third as many as HarnessVLN.

With Qwen3.8-27B~\cite{qwen2026qwen3827b}, case retrieval and four RSI revisions raise cumulative SR on OpenNav-100 from 58\% to 81\%. Stopping is a major source of error: on RecurSet-200, RecurNav with Qwen3-VL-8B~\cite{bai2025qwen3vl} reaches the goal region on 72 routes but stops successfully on only 15.

\begin{table}[!htbp]
\centering
\caption{Experience reuse within one IR2R-CE~\cite{krantz2023iterative} tour (RecurNav with GPT-6.1-sol).}
\label{tab:ir2r-recovery}
\footnotesize
\setlength{\tabcolsep}{2pt}
\begin{tabular*}{\linewidth}{@{\extracolsep{\fill}}lrrrrr@{}}
\toprule
Setting & Solved & SR$\uparrow$ & SPL$\uparrow$ & nDTW$\uparrow$ & NE (m)$\downarrow$\\
\midrule
First pass & 31/47 & 65.96 & 61.91 & 73.41 & 3.04\\
+ Experience & \textbf{35/47} & \textbf{74.47} & \textbf{69.59} & \textbf{76.27} & \textbf{2.72}\\
\bottomrule
\end{tabular*}
\end{table}

\subsection{Navigation Efficiency}
\begin{figure}[!htbp]
\centering
\includegraphics[width=\linewidth]{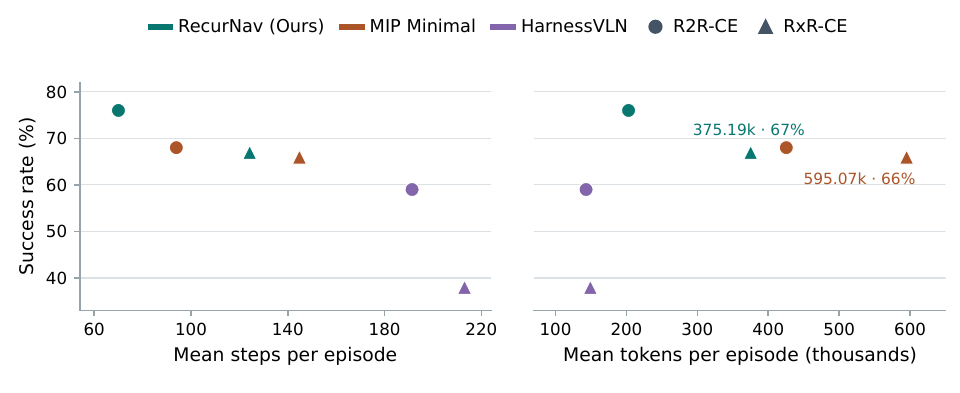}
\caption{Success rate versus mean steps (left) and mean tokens (right) per episode. Token counts include cached reads.}
\label{fig:efficiency-per-episode}
\end{figure}

Route memory keeps the visual input compact. An RxR-CE decision draws on 8.9 views on average, although the agent has already seen 92.3. On R2R-CE, the corresponding averages are 8.8 and 60.7 views. Over 100 RxR-CE episodes, RecurNav reads 37.2M input tokens and generates 0.35M, whereas MIP Minimal reads 59.3M and generates 0.17M. In total, RecurNav uses 37\% fewer tokens (37.5M vs.\ 59.5M) at a similar success rate (Figure~\ref{fig:efficiency-per-episode}).

\subsection{Experience-Guided Recovery and Transfer}
\begin{figure}[!htb]
\centering
\includegraphics[width=\columnwidth]{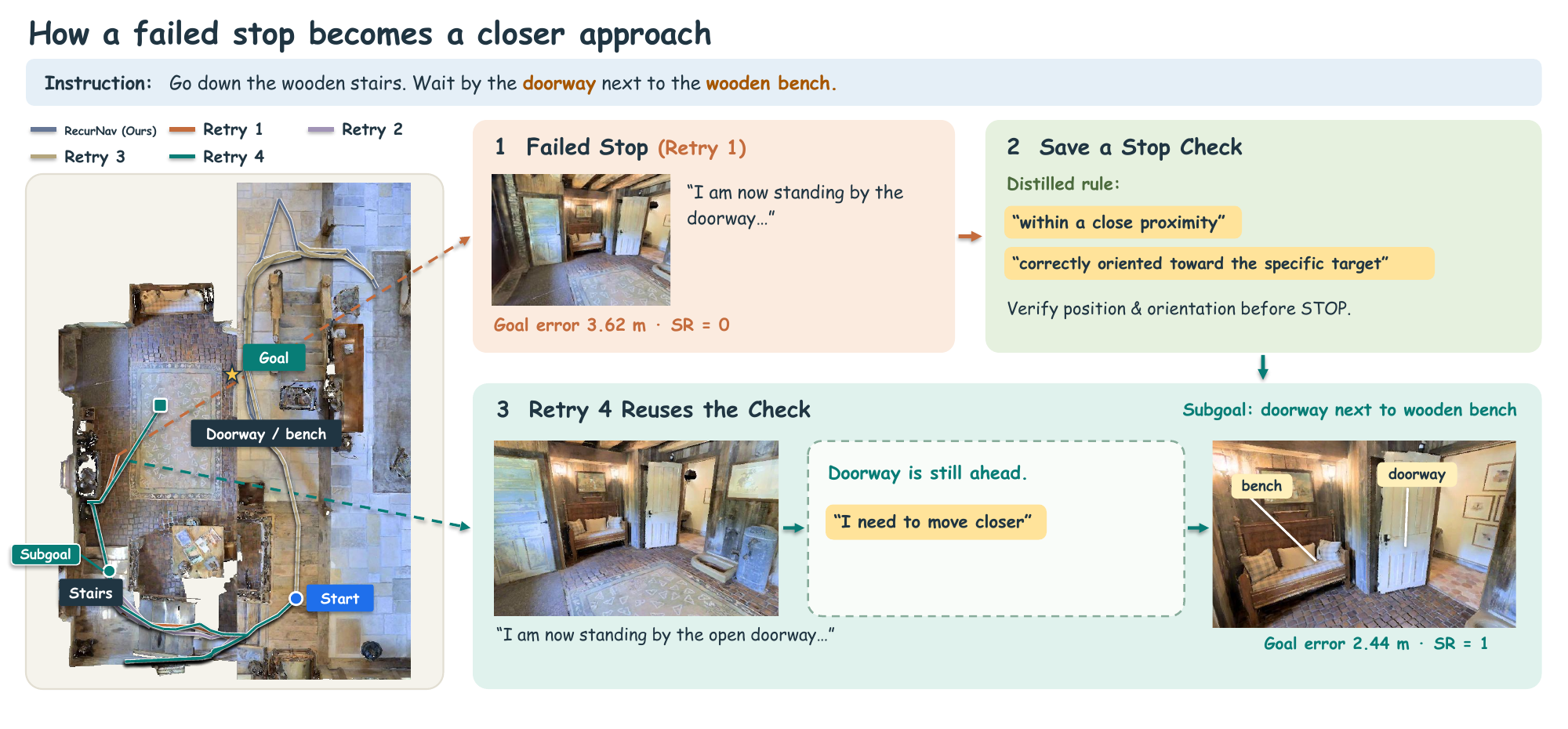}
\caption{A failed stop becomes a closer approach. On retry~1 (historical cases only), RecurNav stops 3.62\,m from the goal. On retry~4, the retrieved check on proximity and orientation leads it to approach the bench-side doorway and stop 2.44\,m from the goal.}
\label{fig:rsi-stop-cases}
\end{figure}

On OpenNav-100, RecurNav with Qwen3.8-27B reaches 85\% cumulative SR after ten retry rounds, up from 58\% on the first attempt (Figure~\ref{fig:rsi-rounds}). The first retry uses historical cases, and subsequent rounds add RSI lessons to guide subgoal completion and stopping.

The benefit also appears with ST-VLN using Qwen3-VL-8B (Figure~\ref{fig:rsi-rounds}). With evolving memory disabled, rerunning 105 episodes reproduces the original outcomes and all eight recorded metrics exactly. Adding our RSI procedure enables ST-VLN to recover further routes.

To test whether experience helps on new routes with a different model, we give Gemini-3.8-Flash-Low the experience collected by Qwen3.8-27B on OpenNav-100. On 100 other R2R-CE val-unseen routes, SR increases from 57\% to 60\% and SPL from 51.61\% to 53.11\% (Table~\ref{tab:experience-transfer}). 

We also evaluate experience reuse within one environment using an IR2R-CE tour of 47 instructions. RecurNav with GPT-6.1-sol completes 31 instructions on the first pass. Reusing that experience on the 16 unresolved instructions recovers four more, raising cumulative SR from 65.96\% to 74.47\% and SPL from 61.91\% to 69.59\% (Table~\ref{tab:ir2r-recovery}).

\subsection{Case Study}
Figure~\ref{fig:rsi-stop-cases} shows how one lesson changes a stopping decision. On retry~1, RecurNav recognizes the bench-side doorway but stops 3.62\,m from the goal, outside the 3\,m success radius. The distilled lesson requires checking both distance and orientation before stopping. On retry~4, the navigator again proposes to stop with the doorway still ahead. The verifier retrieves the check, and the agent moves closer and stops 2.44\,m from the goal.

\section{Conclusion}
We presented RecurNav, a training-free VLN harness that lets a frozen VLM judge route progress from memory. Spatio-temporal memory and world-coordinate waypoints let the navigator check its progress and correct its course mid-move, while recursive self-improvement turns failed attempts into checks that the verifier applies before accepting a subgoal or a stop. RecurNav reaches 76\% SR on OpenNav-100 with GPT-6.1-sol, and experience-guided retries raise the cumulative SR of Qwen3.8-27B on OpenNav-100 from 58\% to 85\%. Stopping remains the main bottleneck, and the retry protocol assumes that a route can be restarted from its original pose, which rarely holds in deployment. Carrying lessons across environments without repeating the same route is a key direction for future work.

\section*{LLM Usage Statement}
We used OpenAI Codex to assist with manuscript preparation.

{
    \small
    \bibliographystyle{ieeenat_fullname}
    \bibliography{main}

@String(CVPR= {IEEE Conf. Comput. Vis. Pattern Recog.})

@String(ECCV= {Eur. Conf. Comput. Vis.})

@String(ICLR = {Int. Conf. Learn. Represent.})

@String(AAAI = {AAAI})

@String(CVPR  = {CVPR})

@String(ECCV  = {ECCV})

@String(ICLR  = {ICLR})

@article{zhou2026mip,
  author = {Jian Zhou and Xunyi Zhao and Gengze Zhou and Zerui Li and Sihao Lin and Jiajun Liu and Qi Wu},
  title = {Embodied Agents Take Control: Minimal-Interface Zero-Shot Agents Rival Industrial-Scale Policies in Vision-and-Language Navigation},
  journal = {arXiv preprint arXiv:2607.26148},
  year = {2026}
}

@article{chen2026harnessvln,
  author = {Yang Chen and Lirong Che and Zhenyu Huang and Wenbo Fu and Chuang Wang and Xu Cao and Daqi Liu and Yuzhe Yang and Jian Su and Lan-Zhe Guo},
  title = {{HarnessVLN: Unifying Training-Free Embodied Navigation through an Agent Harness}},
  journal = {arXiv preprint arXiv:2609.15195},
  year = {2026}
}

@article{zhao2026navharness,
  author = {Xunyi Zhao and Jian Zhou and Sihao Lin and Gengze Zhou and Zerui Li and Xinyu Yan and Jiajun Liu and Anton van den Hengel and Qi Wu},
  title = {{NavHarness: Towards Lifelong Embodied Navigation}},
  journal = {arXiv preprint arXiv:2609.34276},
  year = {2026}
}

@inproceedings{anderson2018r2r,
  author = {Peter Anderson and Qi Wu and Damien Teney and Jake Bruce and Mark Johnson and Niko S{\"u}nderhauf and Ian Reid and Stephen Gould and Anton van den Hengel},
  title = {{Vision-and-Language Navigation: Interpreting Visually-Grounded Navigation Instructions in Real Environments}},
  booktitle = {Proceedings of the IEEE Conference on Computer Vision and Pattern Recognition ({CVPR})},
  pages = {3674--3683},
  month = {June},
  year = {2018}
}

@inproceedings{ku2020rxr,
  author = {Alexander Ku and Peter Anderson and Roma Patel and Eugene Ie and Jason Baldridge},
  title = {{Room-Across-Room: Multilingual Vision-and-Language Navigation with Dense Spatiotemporal Grounding}},
  booktitle = {Proceedings of the 2020 Conference on Empirical Methods in Natural Language Processing ({EMNLP})},
  pages = {4392--4412},
  month = {November},
  year = {2020},
  publisher = {Association for Computational Linguistics},
  doi = {10.18653/v1/2020.emnlp-main.356}
}

@inproceedings{krantz2020continuous,
  author = {Jacob Krantz and Erik Wijmans and Arjun Majumdar and Dhruv Batra and Stefan Lee},
  title = {{Beyond the Nav-Graph: Vision-and-Language Navigation in Continuous Environments}},
  booktitle = {Computer Vision -- {ECCV} 2020},
  series = {Lecture Notes in Computer Science},
  volume = {12373},
  pages = {104--120},
  year = {2020},
  publisher = {Springer},
  doi = {10.1007/978-3-030-58604-1_7}
}

@inproceedings{wang2026progressthink,
  author = {Shuo Wang and Yucheng Wang and Guoxin Lian and Yongcai Wang and Maiyue Chen and Kaihui Wang and Bo Zhang and Zhizhong Su and Yutian Zhou and Wanting Li and Deying Li and Zhaoxin Fan},
  title = {{Progress-Think}: Semantic Progress Reasoning for Vision-Language Navigation},
  booktitle = {Proceedings of the IEEE/CVF Conference on Computer Vision and Pattern Recognition ({CVPR})},
  pages = {4076--4086},
  month = {June},
  year = {2026}
}

@inproceedings{xue2026profocus,
  author = {Wei Xue and Mingcheng Li and Xuecheng Wu and Jingqun Tang and Dingkang Yang and Lihua Zhang},
  title = {{ProFocus}: Proactive Perception and Focused Reasoning in Vision-and-Language Navigation},
  booktitle = {Proceedings of the IEEE/CVF Conference on Computer Vision and Pattern Recognition ({CVPR})},
  pages = {18129--18139},
  month = {June},
  year = {2026}
}

@inproceedings{zuo2026fantasyvln,
  author = {Jing Zuo and Lingzhou Mu and Fan Jiang and Chengcheng Ma and Mu Xu and Yonggang Qi},
  title = {{FantasyVLN}: Unified Multimodal Chain-of-Thought Reasoning for Vision-and-Language Navigation},
  booktitle = {Proceedings of the IEEE/CVF Conference on Computer Vision and Pattern Recognition ({CVPR})},
  pages = {29674--29683},
  month = {June},
  year = {2026}
}

@inproceedings{dai2026evonav,
  author = {Guangzhao Dai and Shuo Wang and Zihan Wang and Guo-Sen Xie and Yang Yang and Jinshan Pan and Qianru Sun and Xiangbo Shu},
  title = {History to Future: Evolving Agent with Experience and Thought for Zero-shot Vision-and-Language Navigation},
  booktitle = {Proceedings of the IEEE/CVF Conference on Computer Vision and Pattern Recognition ({CVPR})},
  pages = {15177--15187},
  month = {June},
  year = {2026}
}

@inproceedings{guo2026awarevln,
  author = {Wenxuan Guo and Xiuwei Xu and Yichen Liu and Xiangyu Li and Hang Yin and Huangxing Chen and Wenzhao Zheng and Jianjiang Feng and Jie Zhou and Jiwen Lu},
  title = {{AwareVLN}: Reasoning with Self-awareness for Vision-Language Navigation},
  booktitle = {Proceedings of the IEEE/CVF Conference on Computer Vision and Pattern Recognition ({CVPR})},
  pages = {4065--4075},
  month = {June},
  year = {2026}
}

@inproceedings{li2026hsgm,
  author = {Kailing Li and Tianwen Qian and Lijin Yang and Yuqian Fu and Jingyu Gong and Xiaoling Wang and Liang He},
  title = {Bridging the {2D-3D} Gap: A Hierarchical Semantic-Geometric Map for Vision Language Navigation},
  booktitle = {Proceedings of the IEEE/CVF Conference on Computer Vision and Pattern Recognition ({CVPR})},
  pages = {15243--15252},
  month = {June},
  year = {2026}
}

@inproceedings{li2026slow4fast,
  author = {Yang Li and Aming Wu and Zihao Zhang and Yahong Han},
  title = {Towards Open Environments and Instructions: General Vision-Language Navigation via Fast-Slow Interactive Reasoning},
  booktitle = {Proceedings of the IEEE/CVF Conference on Computer Vision and Pattern Recognition ({CVPR})},
  pages = {25184--25192},
  month = {June},
  year = {2026}
}

@inproceedings{shinn2023reflexion,
  author = {Noah Shinn and Federico Cassano and Beck Labash and Ashwin Gopinath and Karthik Narasimhan and Shunyu Yao},
  title = {Reflexion: Language Agents with Verbal Reinforcement Learning},
  booktitle = {Advances in Neural Information Processing Systems},
  volume = {36},
  pages = {8634--8652},
  year = {2023},
  doi = {10.52202/075280-0377}
}

@article{peng2026remember,
  author = {Yuhang Peng and Zhikai Wang and Ziyi Wang and Jiaru Zhong and Xiaoji Zheng and Jingzhi Huang and Yi Wang and Jiangtao Gong},
  title = {What Should a Navigator Remember? Spatiotemporal Memory for Zero-Shot Vision-Language Navigation},
  journal = {arXiv preprint arXiv:2610.38318},
  year = {2026}
}

@inproceedings{zhang2024navid,
  author = {Jiazhao Zhang and Kunyu Wang and Rongtao Xu and Gengze Zhou and Yicong Hong and Xiaomeng Fang and Qi Wu and Zhizheng Zhang and He Wang},
  title = {NaVid: Video-based VLM Plans the Next Step for Vision-and-Language Navigation},
  booktitle = {Proceedings of Robotics: Science and Systems ({RSS})},
  year = {2024},
  doi = {10.15607/RSS.2024.XX.079},
  url = {https://www.roboticsproceedings.org/rss20/p079.html}
}

@article{zhang2025mapnav,
  author = {Lingfeng Zhang and Xiaoshuai Hao and Qinwen Xu and Qiang Zhang and Xinyao Zhang and Pengwei Wang and Jing Zhang and Zhongyuan Wang and Shanghang Zhang and Renjing Xu},
  title = {MapNav: A Novel Memory Representation via Annotated Semantic Maps for Vision-and-Language Navigation},
  journal = {arXiv preprint arXiv:2502.13451},
  year = {2025}
}

@inproceedings{long2024instructnav,
  author = {Yuxing Long and Wenzhe Cai and Hongcheng Wang and Guanqi Zhan and Hao Dong},
  title = {InstructNav: Zero-shot System for Generic Instruction Navigation in Unexplored Environment},
  booktitle = {Proceedings of The 8th Conference on Robot Learning},
  series = {Proceedings of Machine Learning Research},
  volume = {270},
  pages = {2049--2060},
  year = {2025},
  publisher = {PMLR},
  url = {https://proceedings.mlr.press/v270/long25b.html}
}

@article{li2026agenticnav,
  author = {Yijian Li and Changze Li and Han Zheng and Jiyuan Cai and Tong Qin and Ming Yang},
  title = {AgenticNav: Zero-Shot Vision-and-Language Navigation as a Tool-Calling Harness},
  journal = {arXiv preprint arXiv:2606.10577},
  year = {2026}
}

@article{liu2026hamvln,
  author = {An Liu and Bingxi Liu and Hongyu Ding and Yixuan Jiang and Yaran Chen and Fulin Tang and Cong Leng and Hong Zhang and Jian Cheng},
  title = {HAM-VLN: Harnessing Hierarchical Agentic Memory for Zero-Shot Vision-and-Language Navigation},
  journal = {arXiv preprint arXiv:2607.29600},
  year = {2026}
}

@article{wei2026dualvln,
  author = {Meng Wei and Chenyang Wan and Jiaqi Peng and Xiqian Yu and Yuqiang Yang and Delin Feng and Wenzhe Cai and Chenming Zhu and Tai Wang and Jiangmiao Pang and Xihui Liu},
  title = {Ground Slow, Move Fast: A Dual-System Foundation Model for Generalizable Vision-and-Language Navigation},
  journal = {arXiv preprint arXiv:2512.08186},
  year = {2025}
}

@article{wang2026lightnav,
  author = {Shaoan Wang and Aocheng Luo and Fei Huang and Jingyi Xu and Xiaoyang Wang and Yueyu Wang and Qianli Ma and Fan Yang and Ran Mei and Jia Wei and Jiangpeng Hu and Xuhao Liu and Hongming Chen and Yuanbin Shao and Yiyang Lin and Ziliang Li and Liang Pan and Xinhang Liu and Yuntao Ma and Tingxiang Fan},
  title = {LightNav-0: Eliciting VLM Spatial Intelligence for Generalist Embodied Navigation},
  journal = {arXiv preprint arXiv:2608.30935},
  year = {2026}
}

@article{chen2025canav,
  author = {Kehan Chen and Dong An and Yan Huang and Rongtao Xu and Yifei Su and Yonggen Ling and Ian Reid and Liang Wang},
  title = {Constraint-Aware Zero-Shot Vision-Language Navigation in Continuous Environments},
  journal = {IEEE Transactions on Pattern Analysis and Machine Intelligence},
  volume = {47},
  number = {11},
  pages = {10441--10456},
  year = {2025},
  doi = {10.1109/TPAMI.2025.3594204}
}

@inproceedings{yin2025gcvln,
  author = {Hang Yin and Haoyu Wei and Xiuwei Xu and Wenxuan Guo and Jie Zhou and Jiwen Lu},
  title = {GC-VLN: Instruction as Graph Constraints for Training-free Vision-and-Language Navigation},
  booktitle = {Proceedings of The 9th Conference on Robot Learning},
  series = {Proceedings of Machine Learning Research},
  volume = {305},
  pages = {1809--1824},
  year = {2025},
  publisher = {PMLR},
  url = {https://proceedings.mlr.press/v305/yin25a.html}
}

@inproceedings{shi2025smartway,
  author = {Xiangyu Shi and Zerui Li and Wenqi Lyu and Jiatong Xia and Feras Dayoub and Yanyuan Qiao and Qi Wu},
  title = {SmartWay: Enhanced Waypoint Prediction and Backtracking for Zero-Shot Vision-and-Language Navigation},
  booktitle = {2025 IEEE/RSJ International Conference on Intelligent Robots and Systems ({IROS})},
  pages = {16923--16930},
  year = {2025},
  doi = {10.1109/IROS60139.2025.11246147}
}

@inproceedings{qiao2025opennav,
  author = {Yanyuan Qiao and Wenqi Lyu and Hui Wang and Zixu Wang and Zerui Li and Yuan Zhang and Mingkui Tan and Qi Wu},
  title = {Open-Nav: Exploring Zero-Shot Vision-and-Language Navigation in Continuous Environment with Open-Source LLMs},
  booktitle = {2025 IEEE International Conference on Robotics and Automation ({ICRA})},
  pages = {6710--6717},
  year = {2025},
  doi = {10.1109/ICRA55743.2025.11127584}
}

@inproceedings{krantz2023iterative,
  author = {Jacob Krantz and Shurjo Banerjee and Wang Zhu and Jason Corso and Peter Anderson and Stefan Lee and Jesse Thomason},
  title = {Iterative Vision-and-Language Navigation},
  booktitle = {Proceedings of the IEEE/CVF Conference on Computer Vision and Pattern Recognition ({CVPR})},
  pages = {14921--14930},
  month = {June},
  year = {2023}
}

@inproceedings{hong2025gsa,
  author = {Haodong Hong and Yanyuan Qiao and Sen Wang and Jiajun Liu and Qi Wu},
  title = {General Scene Adaptation for Vision-and-Language Navigation},
  booktitle = {The Thirteenth International Conference on Learning Representations ({ICLR})},
  year = {2025},
  url = {https://proceedings.iclr.cc/paper_files/paper/2025/hash/aebf6284fe85a8f44b4785d41bc8249a-Abstract-Conference.html}
}

@article{tu2026retronav,
  author = {Diantao Tu and Yangdong Liu and Niannian Dai and Mengqi Rong and Shuhan Shen},
  title = {{RetroNav}: Scene-Grounded Retrospection and Retry for Self-Evolving Lifelong Navigation},
  journal = {arXiv preprint arXiv:2610.35172},
  year = {2026}
}

@article{yu2026recuris,
  author = {Zhaochen Yu and Yingcheng Wu and Zhenfei Yin and Kaiyuan Chen and Zhe Zhao and Mengdi Wang and Shuicheng Yan and Ling Yang},
  title = {Recursive Experiential-Working Memory Evolution for Long-Horizon Agent Harnesses},
  journal = {arXiv preprint arXiv:2608.24876},
  year = {2026}
}

@article{zhang2026sma,
  author = {Haokai Zhang and Yuhang Ding and Yunshu Zhou and Xinze Du and Shengtao Zhang and Zhiyue Zhao and Yuling Xi and Hao Chen},
  title = {{Spatial Memory Agent}: Experience-Grounded Procedural Memory for Spatial Intelligence},
  journal = {arXiv preprint arXiv:2608.12743},
  year = {2026}
}

@misc{qwen2026qwen3827b,
  author = {{Qwen Team}},
  title = {{Qwen3.8-27B} Model Card},
  year = {2026},
  url = {https://huggingface.co/Qwen/Qwen3.8-27B},
  howpublished = {Hugging Face}
}

@article{bai2025qwen3vl,
  author = {Shuai Bai and Yuxuan Cai and Ruizhe Chen and Keqin Chen and Xionghui Chen and Zesen Cheng and Lianghao Deng and Wei Ding and Chang Gao and Chunjiang Ge and Wenbin Ge and Zhifang Guo and Qidong Huang and Jie Huang and Fei Huang and Binyuan Hui and Shutong Jiang and Zhaohai Li and Mingsheng Li and Mei Li and Kaixin Li and Zicheng Lin and Junyang Lin and Xuejing Liu and Jiawei Liu and Chenglong Liu and Yang Liu and Dayiheng Liu and Shixuan Liu and Dunjie Lu and Ruilin Luo and Chenxu Lv and Rui Men and Lingchen Meng and Xuancheng Ren and Xingzhang Ren and Sibo Song and Yuchong Sun and Jun Tang and Jianhong Tu and Jianqiang Wan and Peng Wang and Pengfei Wang and Qiuyue Wang and Yuxuan Wang and Tianbao Xie and Yiheng Xu and Haiyang Xu and Jin Xu and Zhibo Yang and Mingkun Yang and Jianxin Yang and An Yang and Bowen Yu and Fei Zhang and Hang Zhang and Xi Zhang and Bo Zheng and Humen Zhong and Jingren Zhou and Fan Zhou and Jing Zhou and Yuanzhi Zhu and Ke Zhu},
  title = {{Qwen3-VL} Technical Report},
  journal = {arXiv preprint arXiv:2511.21631},
  year = {2025}
}

@article{huang2026refbeliefnav,
  author = {Jingzhi Huang and Junkai Huang and Yuhang Peng and Keyu Mu and Jiaru Zhong and Yi Wang and Lujia Wang and Lap-Pui Chau and Siyang Gao},
  title = {Have You Seen It Clearly? Uncertainty-Guided Belief Refinement for Zero-Shot Object Navigation},
  journal = {arXiv preprint arXiv:2610.15375},
  year = {2026}
}

@article{huang2026lerrnav,
  author = {Jingzhi Huang and Junkai Huang and Yuhang Peng and Keyu Mu and Hailong Huang and Haoyang Yang and Renbo Luo and Lujia Wang and Yi Wang},
  title = {{LERR-Nav}: Learning to Coordinate Exploration, Recovery, and Reminiscing for Object Navigation},
  journal = {arXiv preprint arXiv:2610.15362},
  year = {2026}
}

@article{peng2025bench2freead,
  author = {Yuhang Peng and Sidong Wang and Jihaoyu Yang and Shilong Li and Han Wang and Jiangtao Gong},
  title = {{Bench2FreeAD}: A Benchmark for Vision-based End-to-end Navigation in Unstructured Robotic Environments},
  journal = {arXiv preprint arXiv:2503.12180},
  year = {2025}
}

@inproceedings{peng2026freeaskworld,
  author = {Yuhang Peng and Yizhou Pan and Xinning He and Jihaoyu Yang and Xinyu Yin and Han Wang and Xiaoji Zheng and Chao Gao and Jiangtao Gong},
  title = {{FreeAskWorld}: An Interactive and Closed-Loop Simulator for Human-Centric Embodied {AI}},
  booktitle = {Proceedings of the AAAI Conference on Artificial Intelligence},
  volume = {40},
  number = {2},
  pages = {926--934},
  year = {2026},
  doi = {10.1609/aaai.v40i2.37061}
}
}

\appendix
\twocolumn[{\centering\Large\bfseries Appendix\par\vspace{1em}}]
\section{Memory Component Ablation}
\label{sec:supp-memory-ablation}

We tested each memory configuration on the same 30 R2R-CE val-unseen routes with Gemini-3.8-Flash-Low (Table~\ref{tab:supp-memory-components}). We sampled the routes by first-pass outcome, drawing 21 successes and nine failures from the full RecurNav run. The navigator used no cross-episode experience or RSI in this comparison.

Task memory lets the navigator maintain a plan and update its progress. Temporal memory gives it access to earlier views, events, and dialogue from previous poses. Spatial history records the route and lets the navigator check for loops and revisit a location. We enabled these components individually and in pairs. Every configuration kept the instruction, current RGB-D observations, \texttt{Preview} views from the current pose, and the local map and motion controller used to execute waypoints.

\begin{table}[t]
\centering
\caption{Gemini-3.8-Flash-Low on 30 R2R-CE val-unseen routes with different combinations of memory. All metrics are percentages.}
\label{tab:supp-memory-components}
\footnotesize
\setlength{\tabcolsep}{3pt}
\begin{tabular*}{\columnwidth}{@{\extracolsep{\fill}}lrrr@{}}
\toprule
Memory available & SR $\uparrow$ & SPL $\uparrow$ & nDTW $\uparrow$ \\
\midrule
Task & 66.67 & 60.66 & 69.07 \\
Temporal & 66.67 & 62.51 & 70.99 \\
Spatial Memory & 56.67 & 50.60 & 65.16 \\
Task + Temporal & 60.00 & 55.22 & 68.61 \\
Task + Spatial Memory & 60.00 & 55.04 & 69.93 \\
Temporal + Spatial Memory & 60.00 & 55.47 & 70.94 \\
\midrule
Full RecurNav & \textbf{70.00} & \textbf{64.44} & \textbf{72.59} \\
\bottomrule
\end{tabular*}
\par\smallskip
{\raggedright\scriptsize Each ablation allowed 160 actions.\par}
\end{table}

Temporal memory alone brought Gemini within one success of the full model (20 of 30 goals versus 21). Task memory alone also reached 20 goals, though its routes scored lower on both SPL and nDTW. Gemini reached 17 goals with spatial history alone and 18 with each two-component combination. The full model scored highest on all three metrics.

\section{Run-to-Run Variability of RSI}
\label{sec:supp-rsi-variability}

Qwen3.8-27B solved 58 of the 100 OpenNav-100 episodes on its first pass. We used the remaining 42 to test how much recovery varies across runs of recursive self-improvement (RSI). Three independent runs began with these same failures and the same bank of historical cases.

At the start of each of the five rounds, the model distilled lessons from the latest failed trajectories, reviewed them, and added the accepted lessons to its bank. It then restarted those episodes from their original poses. Once an episode succeeded, it left the retry pool, but its lessons remained available for later rounds. The next independent run started again with the historical cases and an empty lesson bank.

The three runs used navigation environment seeds 3407, 3408, and 3409 with the same model sampling settings. Each retry allowed 250 actions. We served Qwen3.8-27B in BF16 with a 16,384-token context limit, using eight inference replicas and two simulator workers per GPU.

\begin{table}[t]
\centering
\caption{Cumulative success rate (\%) on OpenNav-100. Each episode counts from its first success. The last column gives the mean and sample standard deviation across the three runs.}
\label{tab:supp-rsi-variability}
\footnotesize
\setlength{\tabcolsep}{2.5pt}
\begin{tabular*}{\columnwidth}{@{\extracolsep{\fill}}lrrrr@{}}
\toprule
Round & Run 1 & Run 2 & Run 3 & Mean $\pm$ SD \\
\midrule
First pass & 58 & 58 & 58 & $58.00\pm0.00$ \\
RSI 1 & 70 & 65 & 64 & $66.33\pm3.21$ \\
RSI 2 & 76 & 68 & 70 & $71.33\pm4.16$ \\
RSI 3 & 78 & 70 & 72 & $73.33\pm4.16$ \\
RSI 4 & 82 & 71 & 72 & $75.00\pm6.08$ \\
RSI 5 & \textbf{85} & \textbf{75} & \textbf{77} & $\mathbf{79.00\pm5.29}$ \\
\bottomrule
\end{tabular*}
\end{table}

The runs recovered 27, 17, and 19 routes, or half of the original failures on average (Table~\ref{tab:supp-rsi-variability}). Cumulative SR reached 85\%, 75\%, and 77\%, respectively. The mean rose in every round and reached 79.00\% after five rounds, with a sample standard deviation of 5.29 percentage points. Section~\ref{sec:supp-matched-retries} compares RSI with a control given the same retry opportunities.

\section{Temporal Memory Design in RecurNav}
\label{sec:supp-temporal-history}

After crossing a doorway, the navigator may need to look back at the view from its approach to decide whether it has entered the intended room. Let $b$ be the first step of the current subgoal and $o_t$ the current observation. An eight-frame window reads $o_{\max(0,t-7):t}$ and loses the starting view $o_b$ once $t-b\geq 8$. Full history retains $o_{0:t}$ in order and packs older views into contact sheets when it reaches the input budget. RecurNav gives priority to the starting and current views, a designated keyframe, and views that record progress on the subgoal. It fills the remaining slots with views sampled by distance traveled. The navigator can then compare the approach and the current view against its measured motion.

We compared all three designs on the same OpenNav-100 routes with Qwen3.8-27B and an empty cross-episode memory at the start of each episode (Table~\ref{tab:supp-temporal-history}). The window and full-history runs shared inference settings. The RecurNav run in the table used a 250-action budget and different sampling settings. A separate RecurNav run with a 160-action budget reached 45\% SR.

\begin{center}
\begin{minipage}{\columnwidth}
\centering
\captionsetup{hypcap=false}
\captionof{table}{Temporal memory on OpenNav-100. SR, SPL, nDTW and OSR are percentages. NE and PL are in meters.}
\label{tab:supp-temporal-history}
\scriptsize
\setlength{\tabcolsep}{1.2pt}
\begin{tabular*}{\linewidth}{@{\extracolsep{\fill}}lrrrrrrr@{}}
\toprule
Memory design & SR $\uparrow$ & SPL $\uparrow$ & nDTW $\uparrow$ & OSR $\uparrow$ & NE $\downarrow$ & PL & Steps \\
\midrule
RecurNav (Ours) & \textbf{58} & \textbf{49.19} & \textbf{61.75} & \textbf{66} & \textbf{5.15} & 12.75 & 98.41 \\
Sliding window & 49 & 42.64 & 57.99 & 62 & 5.39 & 12.06 & 89.39 \\
Full history & 38 & 32.67 & 52.10 & 51 & 6.46 & 11.81 & 92.20 \\
\bottomrule
\end{tabular*}
\end{minipage}
\end{center}

RecurNav completed 22 routes that full history missed, while full history completed two that RecurNav missed. Most of the difference arose on the approach to the goal. RecurNav entered the goal region on 66 routes and full history on 51. After entering the region, RecurNav failed on eight routes and full history on 13. On the 36 routes both completed, their mean SPLs were close (84.4\% and 85.9\%, respectively). Across all 100 routes, RecurNav also followed the reference paths more closely (61.75 versus 52.10 nDTW).

The eight-frame window also reached the goal region more often than full history (62 routes versus 51). Both designs failed on 13 routes after entering the region. The difference in arrivals thus accounts for all 11 extra successes.

\section{Matched Retry Control for Qwen3.8-27B}
\label{sec:supp-matched-retries}

We replayed the retry schedule of a ten-round Qwen3.8-27B run on OpenNav-100 to measure how many routes the model could recover through repeated attempts alone. The original run retrieved historical cases in round~1 and revised its lessons with RSI in the next nine rounds.

The original run solved 58 routes on its first pass and retried the remaining failures after each round. The control ran a fresh first pass and then retried the same episodes in the same order as the original run, including any that the control had already solved. This gave both conditions 234 retries in addition to the initial 100 attempts (Table~\ref{tab:supp-matched-retries}).

Both conditions used the same navigation code, episodes, ground truth, sensors, model weights, and simulator and inference settings. Each allowed 250 actions per attempt and used eight GPUs with two simulator workers per GPU, temperature~0, and seed~3407. The control kept online planning and disabled experience generation, retrieval, and memory-based feedback on completion proposals.

Even with these settings fixed, the fresh first pass solved 49 routes. Of the 58 original successes, 19 became failures, and ten of the 42 original failures became successes. Just one episode matched the original run on all eight recorded navigation metrics.

The shared-baseline columns in Table~\ref{tab:supp-matched-retries} score both sets of retries from the same 58 original successes. The independent control column scores the same control retries from its own 49 first-pass successes. Cumulative SR counts each solved episode once. For shared-baseline SPL, we keep the original scores for the 58 first-pass successes and use the first successful retry for each recovered episode. Unsolved episodes contribute zero. All scores use the full set of 100 routes.

\begin{table}[t]
\centering
\caption{Cumulative SR (\%) when the control follows the ten-round retry schedule of the RSI run. The final row gives cumulative SPL (\%).}
\label{tab:supp-matched-retries}
\footnotesize
\setlength{\tabcolsep}{3pt}
\renewcommand{\arraystretch}{1.08}
\begin{tabular*}{\columnwidth}{@{\extracolsep{\fill}}lrrrr@{}}
\toprule
& & \multicolumn{2}{c}{Shared baseline} & Independent \\
Round & Attempts & Control & RSI & Control \\
\midrule
Baseline & 100 & 58 & 58 & 49 \\
1 & 42 & 67 & 67 & 52 \\
2 & 33 & 70 & 75 & 54 \\
3 & 25 & 71 & 77 & 55 \\
4 & 23 & 71 & 80 & 55 \\
5 & 20 & 72 & 81 & 56 \\
6 & 19 & 73 & 81 & 57 \\
7 & 19 & 74 & 82 & 58 \\
8 & 18 & 75 & 82 & 58 \\
9 & 18 & 76 & 83 & 59 \\
10 & 17 & \textbf{76} & \textbf{85} & \textbf{59} \\
\midrule
Final SPL (\%) & -- & 62.95 & 64.81 & 48.19 \\
\bottomrule
\end{tabular*}
\end{table}

From the shared first pass, both conditions reached 67 successes after one retry round. Over ten rounds, the control recovered 18 of the 42 failures and the run with evolving memory recovered 27. With the same retry schedule, RSI finished nine points higher in cumulative SR and 1.86 points higher in SPL. From its own first pass, the control reached 59\% SR and 48.19\% SPL.

\section{Fixed and Evolving Experience Banks}
\label{sec:supp-fixed-experience}

Starting from the same 75 successes on OpenNav-100, we replayed rounds~3--10 with Qwen3.8-27B and kept the \texttt{rsi-001} experience bank fixed. Both conditions used the same 159 scheduled retries. Updating the bank between rounds recovered four more routes and improved cumulative SPL by 1.22 points (Table~\ref{tab:supp-fixed-experience}).

\noindent\begin{minipage}{\columnwidth}
\centering
\captionsetup{hypcap=false}
\captionof{table}{Fixed and recursively updated experience banks on OpenNav-100. Retries are counted per condition from the shared round~2 starting point. Cumulative SR counts each route once, at its first success.}
\label{tab:supp-fixed-experience}
\footnotesize
\begin{tabular*}{\columnwidth}{@{\extracolsep{\fill}}lrrr@{}}
\toprule
& & \multicolumn{2}{c}{Experience bank} \\
\cmidrule(l){3-4}
Checkpoint & Retries & Fixed & Updated \\
\midrule
Round 2 (shared SR, \%) & 0 & 75 & 75 \\
Round 4 (SR, \%) & 48 & 77 & 80 \\
Round 7 (SR, \%) & 106 & 80 & 82 \\
Round 10 (SR, \%) & 159 & 81 & 85 \\
\midrule
New successes & & 6 & 10 \\
Final SPL (\%) & & 63.59 & 64.81 \\
\bottomrule
\end{tabular*}
\end{minipage}\par\smallskip

\onecolumn
\raggedbottom
\captionsetup{hypcap=false}
\section{Case Studies of Memory-Guided Recovery}
\label{sec:supp-memory-cases}

Table~\ref{tab:supp-case-summary} compares endpoint errors on eight routes. In six cases the gate retrieved a lesson from another episode.

\begin{center}
\begin{minipage}{\dimexpr(\textwidth-\columnsep)/2\relax}
\centering
\captionof{table}{Navigation error (m) on the earlier attempt and the successful retry.}
\label{tab:supp-case-summary}
\small
\setlength{\tabcolsep}{5pt}
\begin{tabular*}{\columnwidth}{@{\extracolsep{\fill}}lrr@{}}
\toprule
Task & Earlier NE & Later NE \\
\midrule
ST-VLN doorway & 9.91 & 0.78 \\
Mantel (1214) & 9.45 & 0.63 \\
Sofa (1266) & 16.63 & 1.77 \\
Sink room (361) & 11.56 & 0.51 \\
Bed (925) & 4.60 & 0.19 \\
Lamp (43) & 10.98 & 0.96 \\
Hallway (785) & 6.65 & 0.56 \\
Table after stairs (229) & 12.58 & 0.51 \\
\bottomrule
\end{tabular*}
\end{minipage}
\end{center}

\noindent\begin{minipage}{\textwidth}
\small
\textbf{Stopping in the next doorway with ST-VLN.}\par\vspace{3pt}
{\centering
\includegraphics[page=2,width=.91\textwidth]{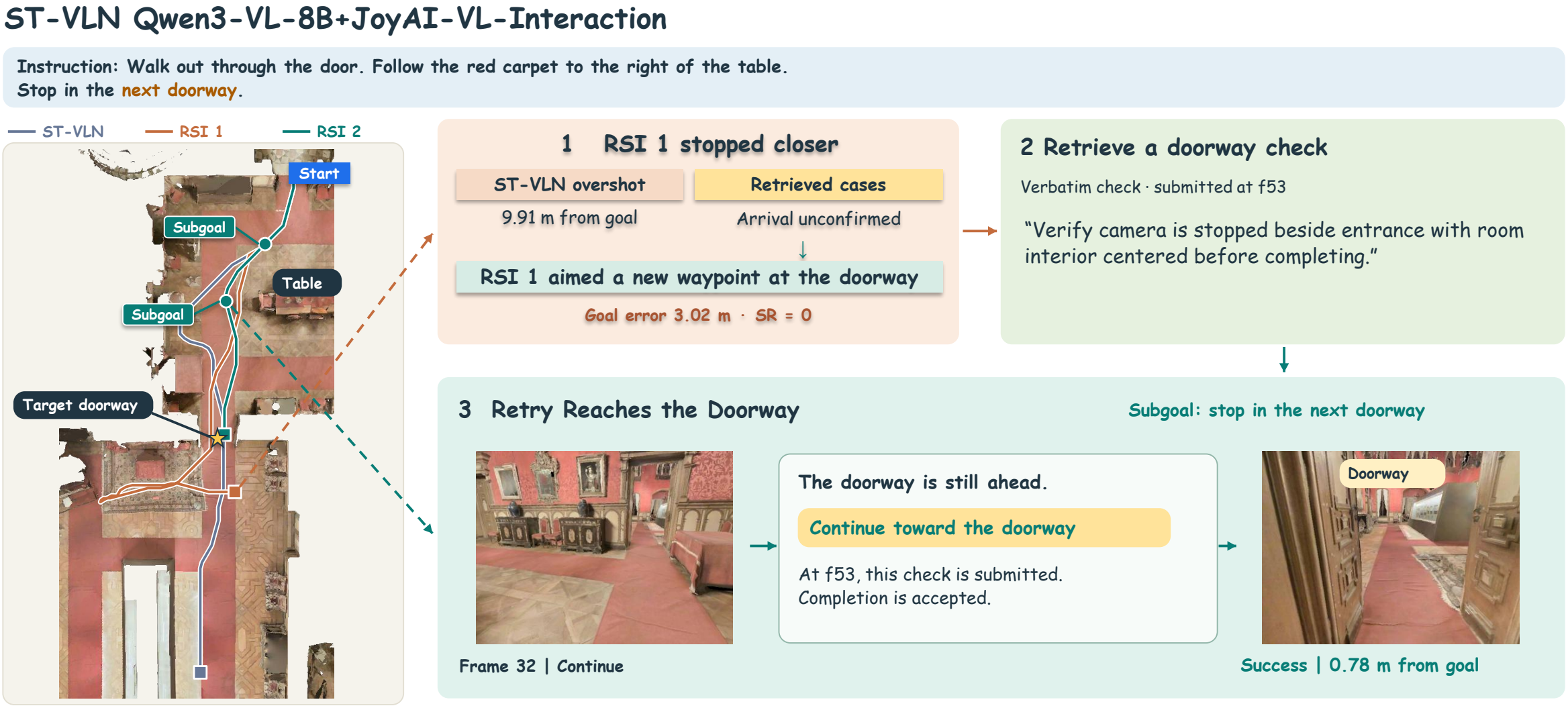}
\captionof{figure}{ST-VLN on the red-carpet route. The doorway is still ahead at frame~32 of the second retry.}
\label{fig:supp-case-stvln}\par}
\vspace{3pt}
The instruction asks the navigator to follow the red carpet to the right of a table and stop in the next doorway (Figure~\ref{fig:supp-case-stvln}). The baseline passed the doorway and finished 9.91\,m from the goal. The first retry aimed a waypoint at the entrance and stopped 3.02\,m from the goal, just outside the 3\,m success radius. At frame~53 of the second retry, the gate received a check asking whether the agent had stopped beside the entrance with the room interior centered in view. It accepted the stop 0.78\,m from the goal.
\end{minipage}\par

\vspace{12pt}
\noindent\begin{minipage}{\textwidth}
\small
\textbf{Stopping beside the door in the sink room.}\par\vspace{3pt}
{\centering
\includegraphics[page=5,width=.91\textwidth]{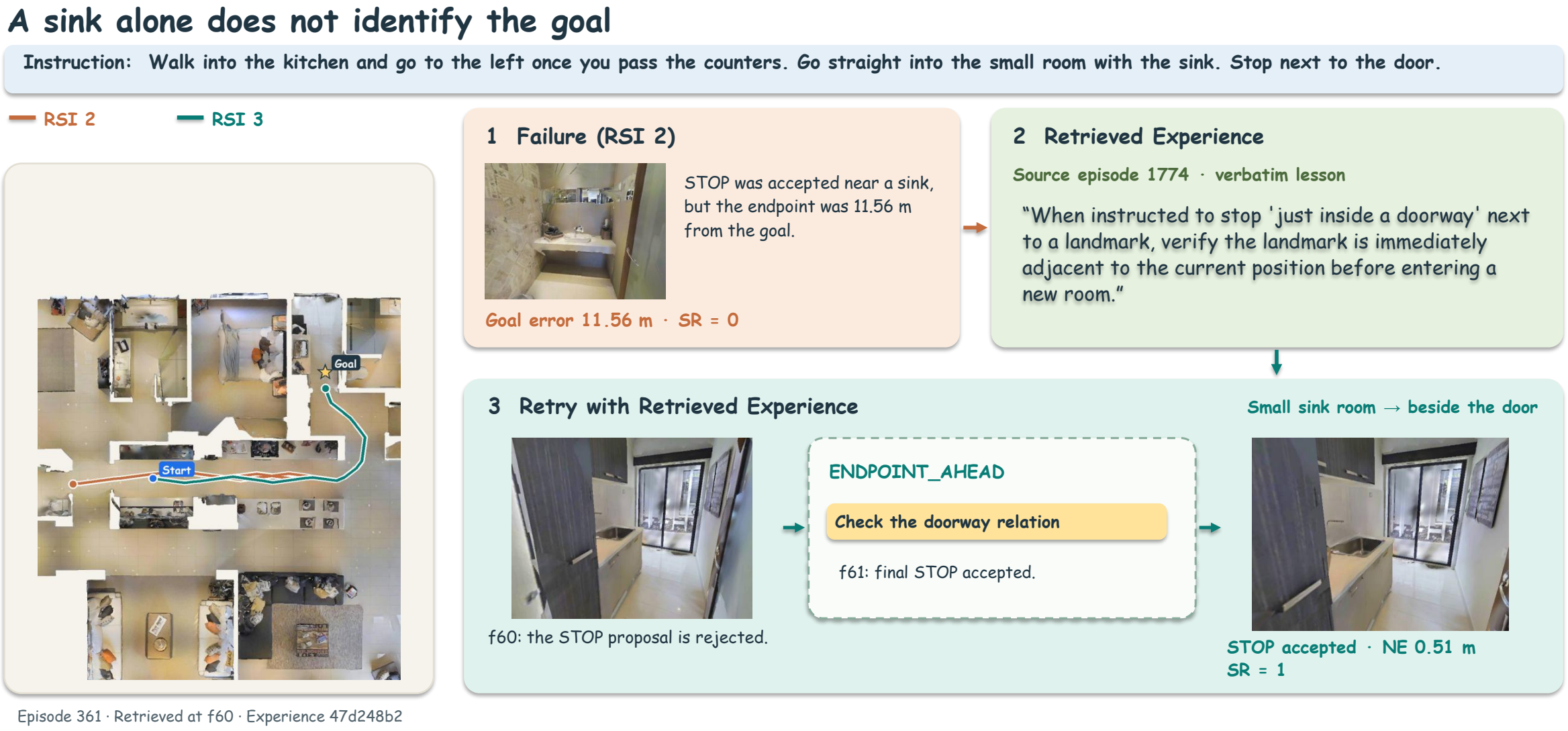}
\captionof{figure}{Episode~361. The earlier trajectory stops near a sink. The instruction specifies the door of the small room beyond the kitchen counters.}
\label{fig:supp-case-sink}\par}
\vspace{3pt}
On the earlier attempt, the agent saw a sink and stopped 11.56\,m from the goal (Figure~\ref{fig:supp-case-sink}). During the retry, the gate retrieved a lesson from episode~1774 that asks whether the named landmark is immediately beside the agent at a doorway. It rejected STOP at frame~60 with \texttt{ENDPOINT\_AHEAD}. One frame later it accepted the stop beside the target door, 0.51\,m from the goal.
\end{minipage}\par

\vspace{12pt}
\noindent\begin{minipage}{\textwidth}
\small
\textbf{Waiting by the bed.}\par\vspace{3pt}
{\centering
\includegraphics[page=6,width=.91\textwidth]{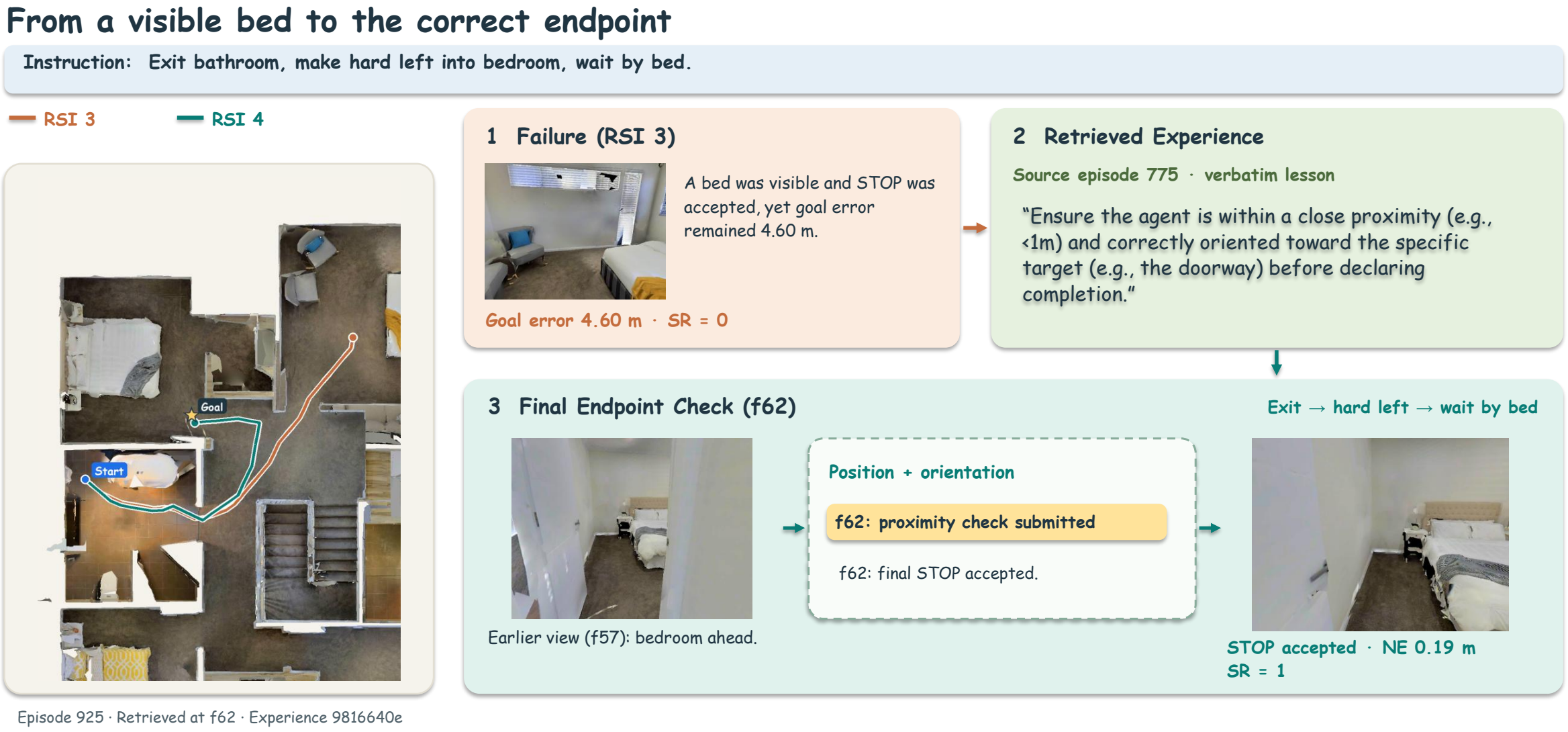}
\captionof{figure}{Episode~925. The agent must leave the bathroom, turn sharply left into the bedroom, and wait beside the bed.}
\label{fig:supp-case-bed}\par}
\vspace{3pt}
A bed was already visible when the gate accepted STOP on the earlier attempt, 4.60\,m from the goal (Figure~\ref{fig:supp-case-bed}). On the retry, frame~57 shows the bedroom ahead. At frame~62 the gate received a lesson from episode~775, also used on the bench route, that asks it to check the agent's orientation and proximity to the target. The lesson suggests a distance of less than 1\,m. The gate accepted STOP beside the bed, 0.19\,m from the goal.
\end{minipage}\par

\vspace{12pt}
\noindent\begin{minipage}{\textwidth}
\small
\textbf{Recovering the branch toward the mantel.}\par\vspace{3pt}
{\centering
\includegraphics[page=3,width=.91\textwidth]{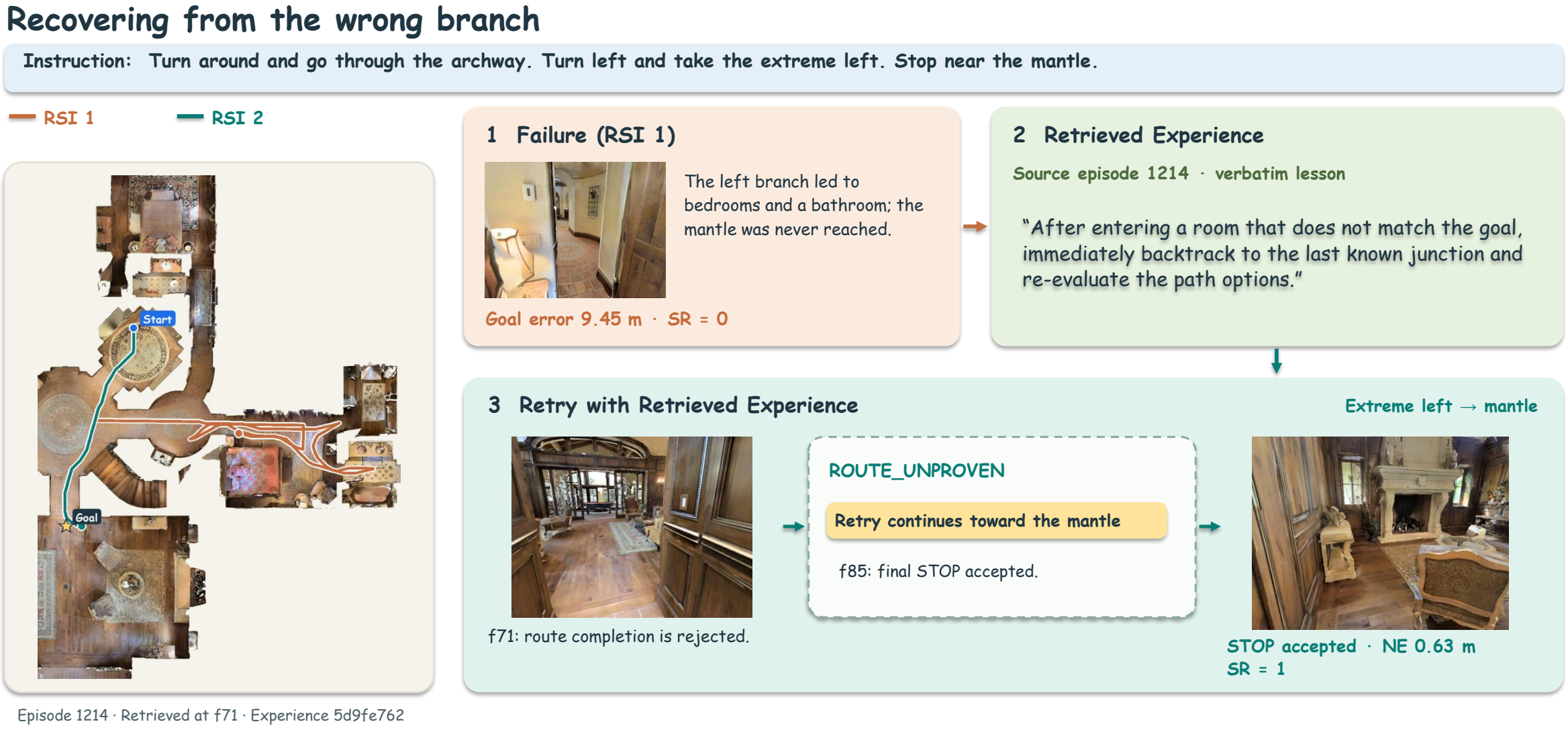}
\captionof{figure}{Episode~1214. The route to the mantel takes the far-left branch after the archway.}
\label{fig:supp-case-mantel}\par}
\vspace{3pt}
The earlier attempt took a branch toward the bedrooms and bathroom and ended 9.45\,m from the goal (Figure~\ref{fig:supp-case-mantel}). A lesson from that same episode advised returning to the last known junction after entering an unexpected room. When the agent claimed at frame~71 that it had taken the far-left branch, the gate retrieved the lesson and returned \texttt{ROUTE\_UNPROVEN}. The agent continued toward the mantel and stopped at frame~85, 0.63\,m from the goal.
\end{minipage}\par

\vspace{12pt}
\noindent\begin{minipage}{\textwidth}
\small
\textbf{Following the hallway to the correct sofa.}\par\vspace{3pt}
{\centering
\includegraphics[page=4,width=.91\textwidth]{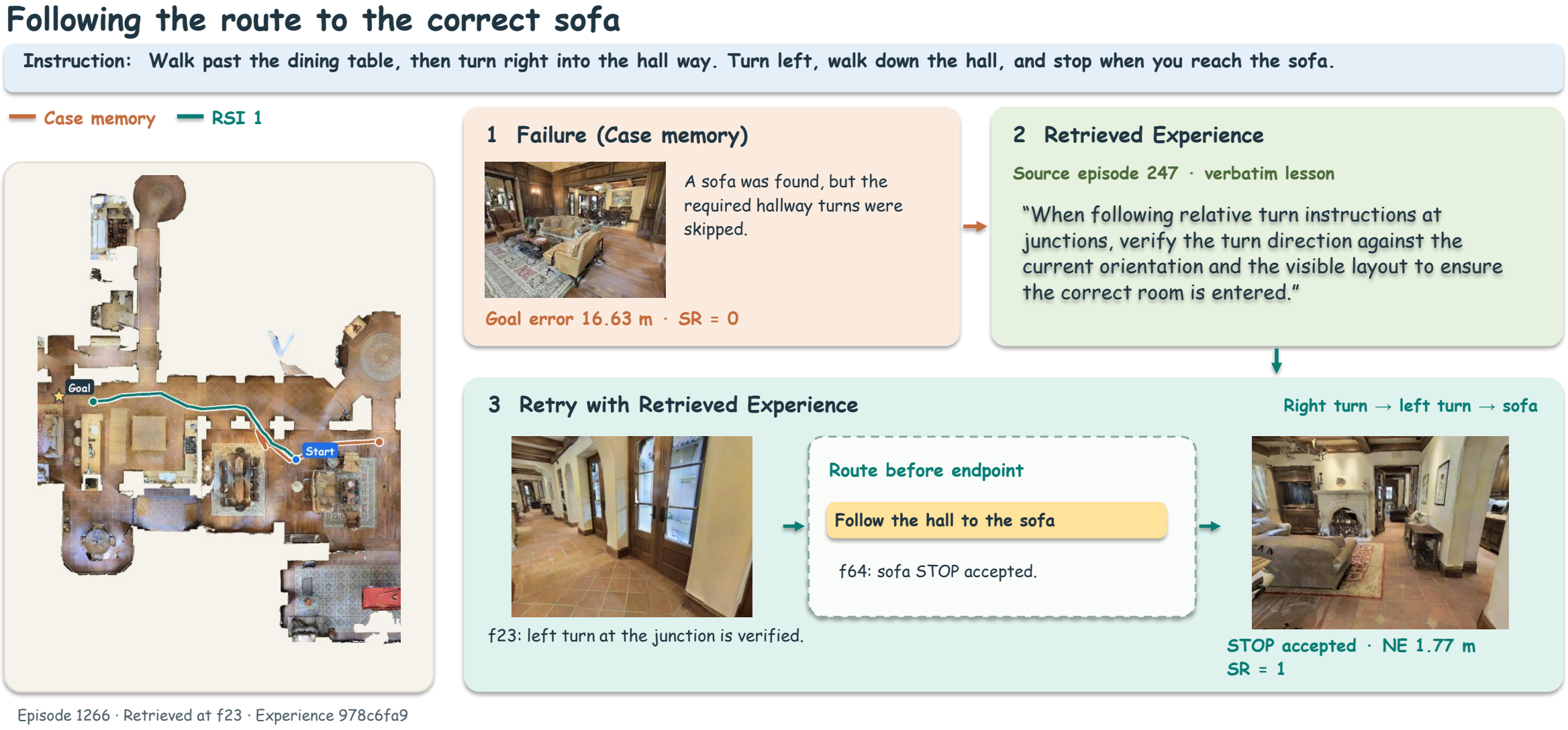}
\captionof{figure}{Episode~1266. A right turn into the hallway followed by a left identifies which sofa the instruction refers to.}
\label{fig:supp-case-sofa}\par}
\vspace{3pt}
The case-only attempt stopped near a sofa 16.63\,m from the goal after skipping the hallway turns (Figure~\ref{fig:supp-case-sofa}). On the retry, a lesson from episode~247 asked the gate to check the turn against the agent's heading and the visible junction. The gate confirmed the left turn at frame~23. The agent followed the corridor to the target sofa and stopped at frame~64, 1.77\,m from the goal.
\end{minipage}\par

\vspace{12pt}
\noindent\begin{minipage}{\textwidth}
\small
\textbf{Finding the lamp beyond the kitchen.}\par\vspace{3pt}
{\centering
\includegraphics[page=7,width=.91\textwidth]{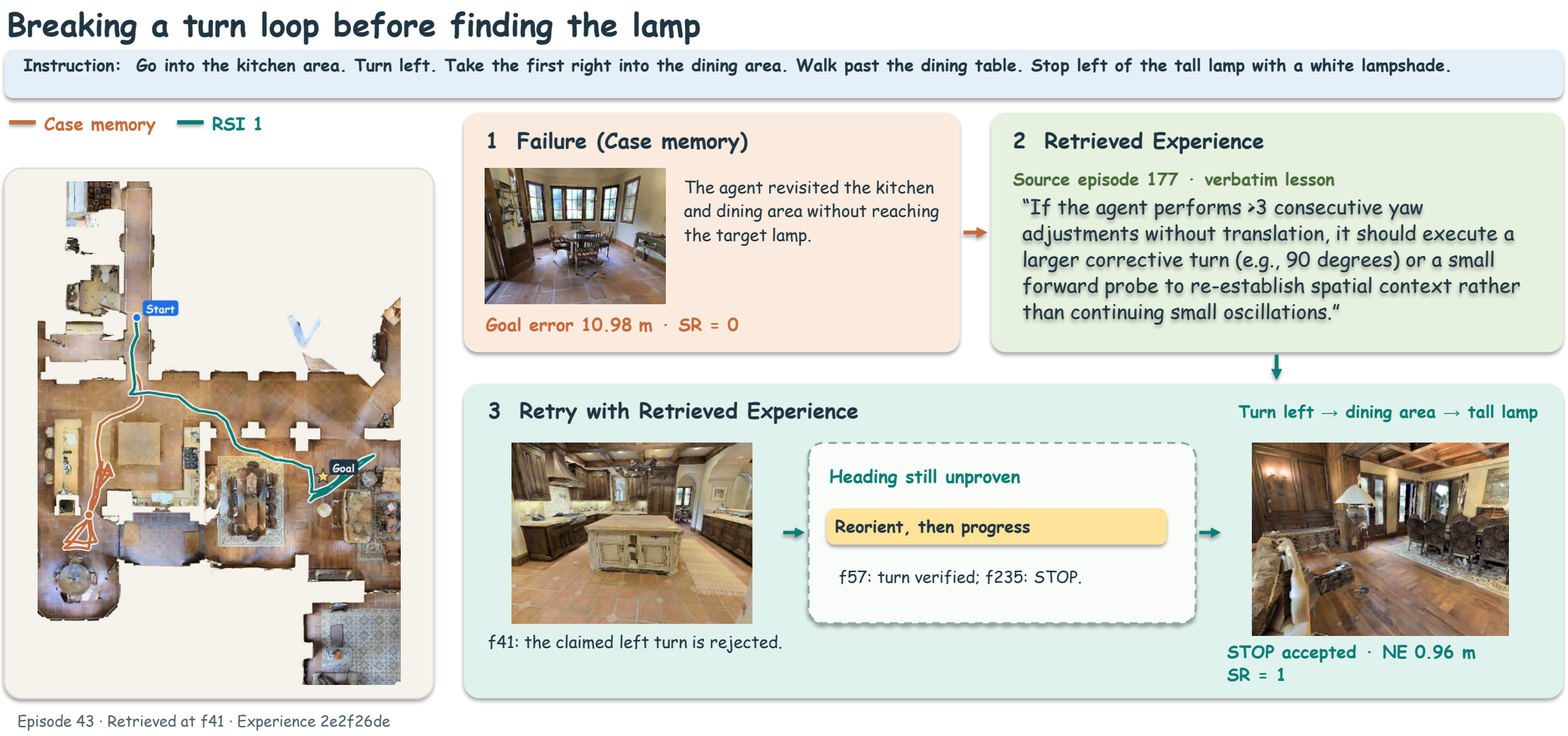}
\captionof{figure}{Episode~43. The case-only trajectory loops between the kitchen and dining area. The retry continues to the tall lamp.}
\label{fig:supp-case-lamp}\par}
\vspace{3pt}
On the case-only attempt, the agent circled between the kitchen and dining area and finished 10.98\,m from the lamp (Figure~\ref{fig:supp-case-lamp}). Episode~177 supplied a lesson for this situation. It recommends a larger turn or a short step forward when the agent keeps turning in place. The gate rejected the claimed kitchen left turn at frame~41 and accepted it at frame~57. The agent reached the lamp and stopped at frame~235, 0.96\,m from the goal.
\end{minipage}\par

\vspace{12pt}
\noindent\begin{minipage}{\textwidth}
\small
\textbf{Leaving the bedroom for the hallway.}\par\vspace{3pt}
{\centering
\includegraphics[page=8,width=.91\textwidth]{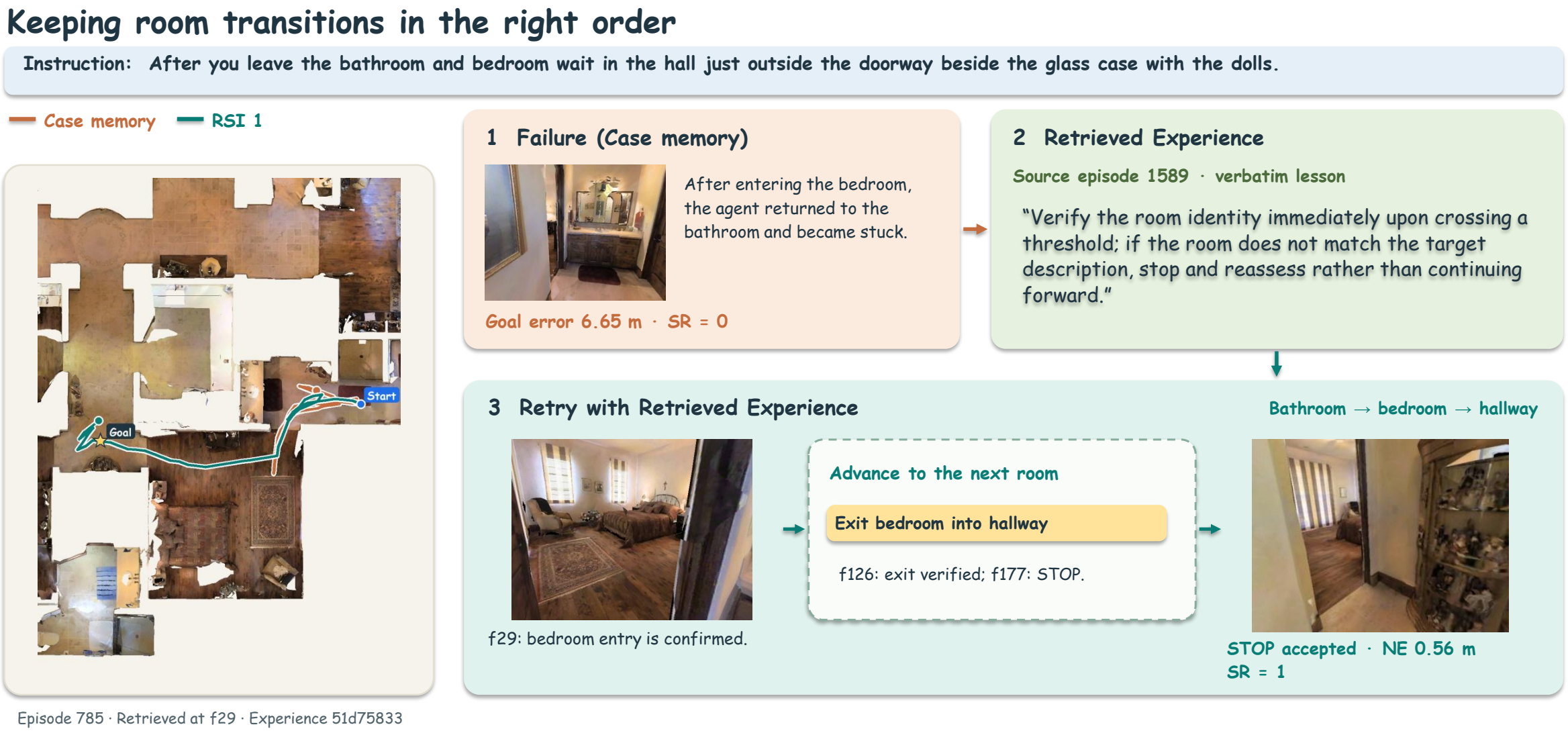}
\captionof{figure}{Episode~785. The goal is in the hallway beside the glass display case, beyond the bathroom and bedroom.}
\label{fig:supp-case-hall}\par}
\vspace{3pt}
After entering the bedroom, the case-only navigator doubled back into the bathroom (Figure~\ref{fig:supp-case-hall}). A lesson from episode~1589 asks the navigator to identify the room it has just entered after crossing a threshold. On the retry, the gate confirmed the bathroom exit at frame~29 and the bedroom exit at frame~126. The agent then stopped in the hallway at frame~177, 0.56\,m from the goal.
\end{minipage}\par

\vspace{12pt}
\noindent\begin{minipage}{\textwidth}
\small
\textbf{Taking the stairs to the table.}\par\vspace{3pt}
{\centering
\includegraphics[page=9,width=.91\textwidth]{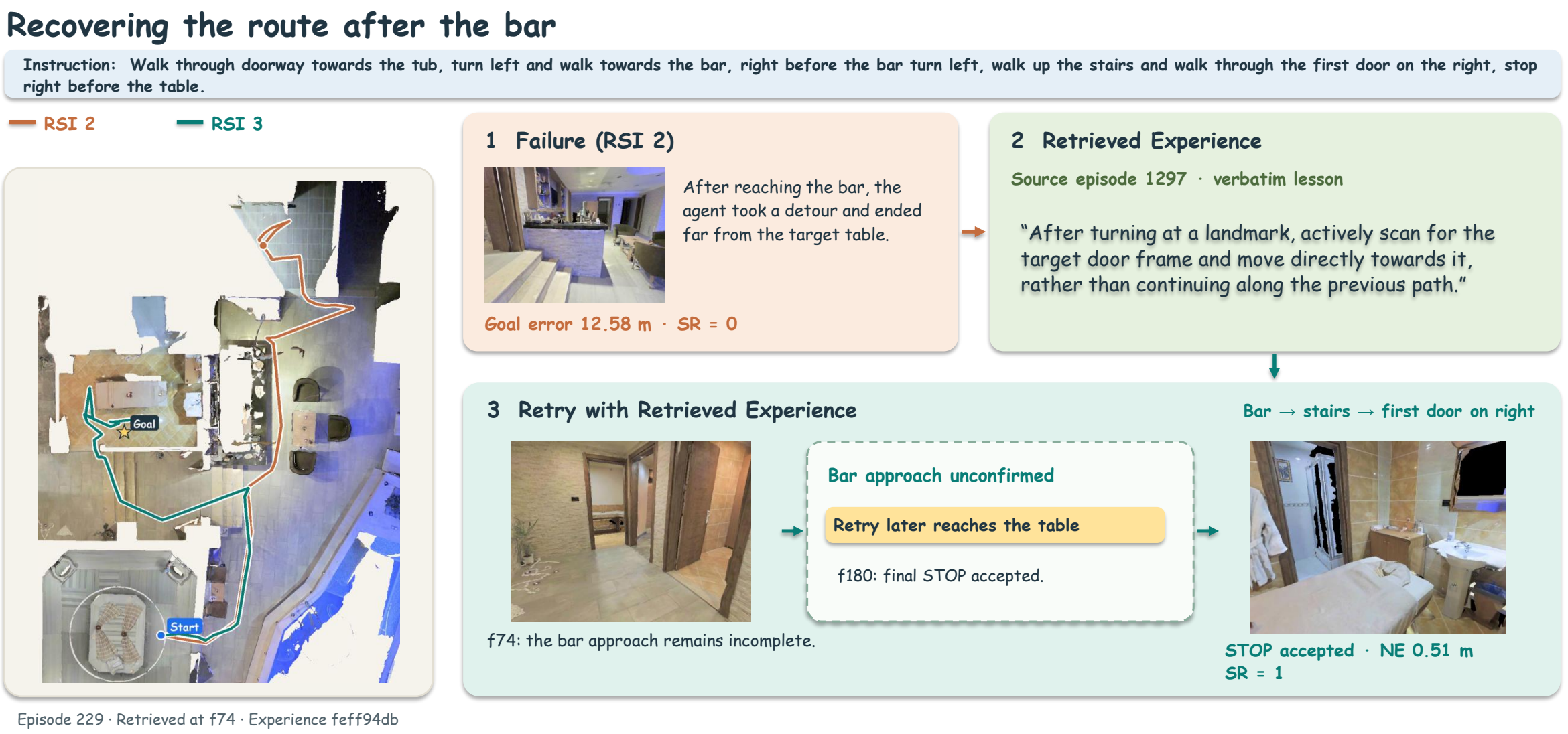}
\captionof{figure}{Episode~229. The instruction calls for a turn before the bar, a climb up the stairs, and a stop by the table beyond the first door on the right.}
\label{fig:supp-case-table}\par}
\vspace{3pt}
The trajectories diverged at the bar, where the earlier attempt missed the approach to the stairs (Figure~\ref{fig:supp-case-table}). At frame~74 of the retry, the gate retrieved a lesson from episode~1297 about looking for the door frame after a turn at a landmark. The bar approach remained open in the plan. Later checks still cited the view from frame~81. At frame~180 the gate accepted the final stop 0.51\,m from the goal while the bar approach still carried the label \texttt{ENDPOINT\_UNPROVEN}.
\end{minipage}\par

\section{Real-World Deployment}
\noindent\begin{minipage}{\textwidth}
\centering
\includegraphics[width=\linewidth]{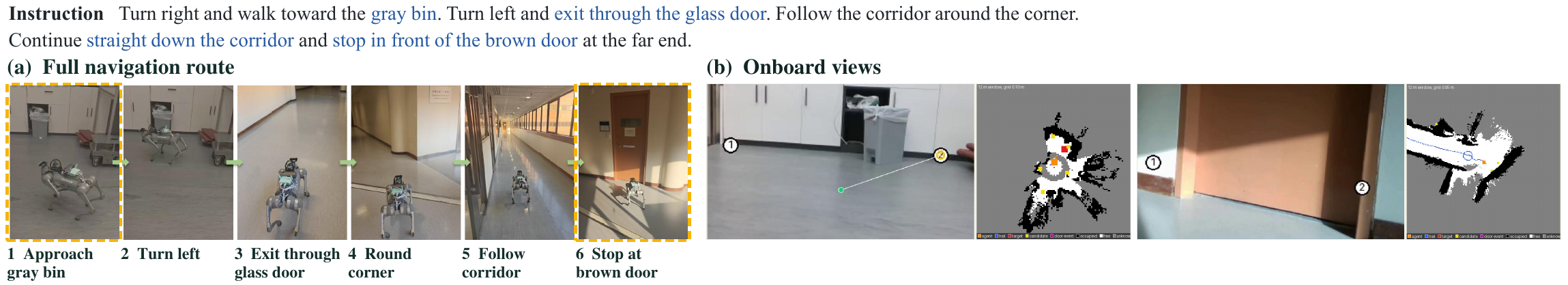}
\captionof{figure}{Real-world deployment on a Unitree Go2. (a) The six stages of the route. (b) Onboard views with numbered waypoint marks and local maps. A remote server runs Qwen3.8-27B and sends motion commands to the robot.}
\label{fig:go2-deployment}
\end{minipage}\par

\end{document}